\documentclass{IEEEtran}
\usepackage{cite}
\usepackage{amsmath,amssymb,amsfonts}
\usepackage{algorithmic}
\usepackage{graphicx}
\usepackage{textcomp}
\usepackage{hyperref}
\usepackage{multirow}
\usepackage{bbm}
\def\BibTeX{{\rm B\kern-.05em{\sc i\kern-.025em b}\kern-.08em
    T\kern-.1667em\lower.7ex\hbox{E}\kern-.125emX}}
\begin{document}

%\bstctlcite{IEEEexample:BSTcontrol}
    \title{Spatiotemporal Forecasting of Traffic Flow using Wavelet-based Temporal Attention}
  \author{Yash Jakhmola, Madhurima Panja, Nitish Kumar Mishra, Kripabandhu Ghosh, Uttam Kumar and~Tanujit Chakraborty
  
\thanks{Corresponding Author: T. Chakraborty is with Sorbonne University, Abu Dhabi and Paris (e-mail: tanujit.chakraborty@sorbonne.ae).}
\thanks{Y. Jakhmola, N. K. Mishra, and K. Ghosh are with Indian Institute of Science, Education and Research, Kolkata, India.}
\thanks{M. Panja and U. Kumar are with International Institute of Information Technology, Bangalore, India}}

%\markboth{IEEE TRANSACTIONS ON RELIABILITY}{Chakraborty \MakeLowercase{\textit{et al.}}: Hellinger Net}

\maketitle

\begin{abstract}
Spatiotemporal forecasting of traffic flow data represents a typical problem in the field of machine learning, impacting urban traffic management systems. In general, spatiotemporal forecasting problems involve complex interactions, nonlinearities, and long-range dependencies due to the interwoven nature of the temporal and spatial dimensions. Due to this, traditional statistical and machine learning methods cannot adequately handle the temporal and spatial dependencies in these complex traffic flow datasets. A prevalent approach in the field combines graph convolutional networks and multi-head attention mechanisms for spatiotemporal processing. This paper proposes a wavelet-based temporal attention model, namely a wavelet-based dynamic spatiotemporal aware graph neural network (W-DSTAGNN), for tackling the traffic forecasting problem. Wavelet decomposition can help by decomposing the signal into components that can be analyzed independently, reducing the impact of non-stationarity and handling long-range dependencies of traffic flow datasets. Benchmark experiments using three popularly used statistical metrics confirm that our proposal efficiently captures spatiotemporal correlations and outperforms ten state-of-the-art models (including both temporal and spatiotemporal benchmarks) on three publicly available traffic datasets. Our proposed ensemble method can better handle dynamic temporal and spatial dependencies and make reliable long-term forecasts. In addition to point forecasts, our proposed model can generate interval forecasts that significantly enhance probabilistic forecasting for traffic datasets.

\end{abstract}

\begin{IEEEkeywords}
Traffic forecasting, Wavelet transformation, Temporal attention, Spatiotemporal data.
\end{IEEEkeywords}

\IEEEpeerreviewmaketitle

%\maketitle

\section{Introduction} \label{sec:introduction}
Rapid urbanization and population growth contribute to severe traffic congestion, which negatively affects both traffic safety and environmental conditions \cite{zheng2019deep}. To address these challenges and mitigate congestion, urban planners are increasingly implementing intelligent transportation systems (ITS) in growing cities and metropolitan areas. Recent advancements in sensing technologies, along with the widespread deployment of ground-based sensors on roads and subways, enable the detection of real-time traffic conditions. The resulting large-scale traffic data collection facilitates the design of early intervention strategies through traffic forecasting \cite{vlahogianni2014short}. These strategies help traffic controllers enhance the efficiency of transportation systems and reduce congestion-related issues.

Accurate traffic flow forecasting, a key component of ITS, has emerged as a prominent research area \cite{zhang2011data}. The development of models capable of predicting and preventing traffic congestion, optimizing traffic regulation, and identifying optimal travel routes is essential for the success of ITS in urban environments. Recently, data-driven traffic flow forecasting methods have gained significant attention, largely due to the availability of real-world datasets like the Performance Measurement System (PeMS) dataset, which is collected from individual sensors deployed across major metropolitan areas of California by California Transportation Agencies (CalTrans).

Previous studies in this domain have primarily focused on forecasting traffic flow by extrapolating historical traffic patterns. Classical time series analysis techniques, ranging from autoregressive integrated moving average (ARIMA) models to multivariate vector autoregression (VAR) models, have been utilized for traffic forecasting tasks \cite{shahriari2020ensemble, chandra2009predictions}. However, these models struggle to accurately capture the complexities of non-stationary time sequences in traffic data. More recently, machine learning approaches like support vector regression (SVR) have been applied to address these challenges \cite{castro2009online}. Despite their benefits, deep learning frameworks—particularly those incorporating attention mechanisms or convolutional layers—have shown superior performance, as they automate preprocessing and can better handle the intricacies of traffic flow data. This has led to the widespread adoption of deep learning architectures in traffic forecasting \cite{lv2014traffic, gao2022short, ma2020multi, zhao2019deep, khan2023short}.

Although deep learning-based temporal architectures have shown promising results, they often fail to capture the spatial dependencies inherent in traffic data effectively \cite{berkani2023spatio, hamdi2022spatiotemporal}. While temporal models primarily focus on learning the historical characteristics to forecast future dynamics, a detailed analysis reveals that both temporal and spatial patterns influence traffic flow \cite{wang2020deep}. For instance, traffic flow is affected by the time of the day we are considering - traffic can be high near residential areas during evening office closure times or near a school during the school opening times. Traffic can also be affected by nearby traffic conditions - if a road has very low traffic (perhaps due to some construction work), the traffic might be higher on nearby roads \cite{xie2020urban}. Moreover, all these patterns can change depending upon the day - traffic near schools and offices will be much less on weekends, while traffic near malls and shopping complexes will be higher on weekends, and vice-versa. These complex patterns make traffic flow prediction a very complicated task \cite{traffic_planning}. To model these dynamic changes in the traffic flow datasets, researchers have focused on designing the problem as a spatiotemporal forecasting setup \cite{ermagun2018spatiotemporal}.  In recent years, graph-based deep learning models, particularly graph neural networks (GNN), have emerged as a powerful tool for handling spatiotemporal datasets in various domains ranging from sensor networks \cite{shi2015infinite, zhu2012graph} and climate modeling \cite{mei2015signal,ray2021optimized} to traffic control systems \cite{choi2022graph, dstagnn, yu2017spatio}. Variants of GNN frameworks have achieved state-of-the-art performance in traffic forecasting problems due to their ability to capture spatial dependencies using non-Euclidean graph structures. For instance, an encoder-decoder architecture (both with spatial and temporal attention) along with a transform attention layer between the encoder and decoder, namely graph multi-attention network (GMAN), has been proposed \cite{gman}. Dynamic spatiotemporal aware graph neural network (DSTAGNN) \cite{dstagnn} has been one of the recent state-of-the-art models for traffic forecasting due to its ability to handle the high-dimensional dynamic spatiotemporal nature of the traffic flow datasets. However, it fails to generate long-term forecasts and capture the seasonal patterns in the temporal structures of the PeMS traffic flow data \cite{pavlyuk2018spatiotemporal}. 

Alongside the forecasting technique, several decomposition techniques, such as Fourier transforms \cite{fourier}, Fast Fourier transforms \cite{fft}, and Wavelet decomposition \cite{waveletbook}, have shown competencies in time series pre-processing tasks, among many others. A recent study by Sasal et al. has shown that when a wavelet-transformed sequence is fed into a transformer (and then inverse-transformed after forecasting), it improves the performance of the transformer for temporal forecasting of long-sequence data \cite{wtransformer}. To overcome the issues with DSTAGNN and other spatiotemporal forecasting models for traffic data, we introduce wavelet-based temporal attention that can effectively model temporal dynamics and spatial patterns of PeMS datasets. Our proposed wavelet-based dynamic spatiotemporal aware graph neural network (W-DSTAGNN) method can simultaneously handle the non-stationarity and nonlinear structure of the spatiotemporal data and can generate long-term forecasts for traffic conditions. In addition, our proposal, combined with conformal prediction, can generate prediction intervals for probabilistic forecasting of traffic flow data, which will be of immense use for traffic management systems. Our contributions can be summarized as follows: 
\begin{enumerate}
    \item We propose a novel framework that combines maximal overlapping discrete wavelet transformation (MODWT) and the temporal attention module as W-DSTAGNN for learning long-term temporal and spatial dependencies of traffic conditions.
    \item The proposed W-DSTAGNN captures the nonlinearity, non-stationarity, and complicated relations between the nodes in a better way than the standard traffic forecasting models. This is confirmed by large-scale experiments with three datasets and eleven baselines. 
    \item Multiple comparisons with the best (MCB) test are performed to show that our model indeed performs better than the baselines. We also presented a conformal prediction plot to give further evidence for the competence of our method in generating prediction intervals.
\end{enumerate}

The rest of the paper is organized as follows: Section \ref{sec:related} reviews various time series forecasting approaches designed for traffic forecasting tasks. Section \ref{sec:maths} briefly describes the features of wavelet decomposition along with its mathematical formulation. Section \ref{sec:methodology} introduces the proposed W-DSTAGNN approach for spatiotemporal forecasting. Section \ref{sec:experiments} highlights the efficiency of the proposal over baseline forecasters using extensive experiments with three real-world datasets and statistical significance tests.  Section \ref{sec:limitations} highlights the drawbacks and limitations of our method along with the scope for future improvements. Finally, Section \ref{sec:conclusion} concludes the paper. 

\section{Related Work}\label{sec:related}
With the number of vehicles on the road increasing every year, it is no surprise that current traffic management systems need to become even more effective. Traffic flow prediction plays a pivotal role in these systems \cite{traffic_planning}; however, accurately predicting traffic flow remains a challenging task. In the past, practitioners and researchers have utilized historical traffic data and explored forecasting approaches from various paradigms to analyze traffic conditions \cite{vlahogianni2014short, lee2021short, tedjopurnomo2020survey}. These studies have focused on modeling both temporal and spatiotemporal dependencies better to understand the complex dynamics within traffic flow datasets. In this section, we provide a brief overview of the various approaches adopted in the literature to address traffic forecasting challenges.

\subsection{Temporal forecasting approaches}

Traditional time series forecasting models have been a popular choice among practitioners for modeling traffic flow datasets \cite{hyndman2018forecasting}. These frameworks typically extrapolate the historical patterns from stationary traffic flow series to predict its future dynamics \cite{petropoulos2022forecasting}. Among these, popular architectures include the linear ARIMA model and its variants, often enhanced with Kalman filtering techniques \cite{arima}. In 2009, Chandra et al. demonstrated how traffic speeds and volumes in Orlando, Florida, were influenced by both upstream and downstream location data and employed a VAR model to predict future traffic conditions \cite{chandra2009predictions}. In recent years, advancements in sensor technologies have led to a significant increase in the availability of traffic flow datasets. To handle this surge of data, data-driven forecasting approaches have become mainstream in traffic prediction. For example, in 2011, Hong et al. \cite{hong2011forecasting} applied a kernel-based SVR model to forecast inter-urban traffic flow in the northern Taiwan region. The main advantage of machine learning techniques over traditional methods lies in their ability to model nonlinear temporal dependencies \cite{chakraborty2019hybrid}. With the improvement of computational capabilities, deep learning architectures have also become an integral part of time series forecasting \cite{lv2014traffic, yu2017deep}. Recurrent neural network-based models, such as long short-term memory (LSTM) networks and their variants, are widely used to capture temporal correlations in traffic flow datasets \cite{fclstm}. These architectures utilize gated mechanisms to regulate information flow, which plays a crucial role in effectively capturing both short-term and long-term dependencies. Although these models offer numerous advantages over conventional forecasting methods, they struggle to incorporate spatial information. The complex spatial dependencies inherent in traffic flow data are difficult to account for using conventional forecasting approaches.

\subsection{Spatiotemporal forecasting approaches}
In recent years, many deep learning techniques have been employed to tackle the problem of high-dimensional spatiotemporal traffic prediction. Convolutional neural networks (CNNs) have been used in traffic forecasting due to their spatial information extraction abilities; for example, \cite{stresnet} converts the road network to a regular 2D grid and applies CNN to predict the flow. Nowadays, graph convolutional networks (GCNs) are used to model spatial correlations in network data \cite{gcn}, which put spectral graph theory into deep neural networks. In another recent work, \cite{chebnet} proposed ChebNet, which boosts GCNs with fast localized convolution filters. More recently, diffusion convolutional recurrent neural network (DCRNN) \cite{dcrnn} introduces graph convolutional networks into spatiotemporal network data prediction, which employs a diffusion graph convolution network to understand the information diffusion process in spatial networks, along with RNN to model temporal correlations. Spatiotemporal synchronous graph convolutional network (STSGCN) \cite{stsgcn} concatenated the spatial graphs of multi-neighborhood time steps. Graph-WaveNet (GWN) \cite{graphwavenet} designed a self-adaptive matrix to understand the changes of the influence between nodes and their neighbors. It used dilated casual convolutions for the temporal correlations, thus increasing the receptive field exponentially. Adaptive graph convolutional recurrent network (AGCRN) \cite{agcrn} found hidden spatial dependencies via learnable embedding from nodes. However, the spatiotemporal layers cannot be stacked to expand the receptive field. GMAN \cite{gman} is an encoder-decoder architecture with spatial and temporal attention modules to model spatiotemporal correlations. It also has a transform attention layer between the encoder and decoder to alleviate error propagation during long-term prediction. Thus, the traffic flow forecasting problem is an emerging research area for both transportation research and machine learning communities working on spatiotemporal data structures. A highly accurate traffic forecasting system impacts our day-to-day life. Our proposed methodology can be a long-term forecasting tool for traffic data modelers. 

\subsection{Wavelet-based forecasters}

Wavelet transformation (WT) has demonstrated remarkable progress in time series analysis by enhancing the efficiency of individual forecasting methods \cite{joo2015time, liu2017time}. WT is particularly useful for extracting signals from noise in the time-frequency domain \cite{waveletbook}. This method decomposes a time series into high-frequency signals, which depict the details or short-term fluctuations, and low-frequency components, which capture smooth long-term trends. This time-frequency localization has made WT a valuable tool in forecasting across diverse fields, including epidemiology \cite{panja2023ensemble}, economics \cite{sengupta2023forecasting}, environmental studies \cite{borah2024wavecatboost}, geophysics \cite{grinsted2004application}, traffic forecasting \cite{jiang2005dynamic}, and others. In traffic forecasting, WT has been employed to remove the noise from the data, allowing for the modeling of the remaining stationary components using methods like the Kalman filter and neural networks \cite{xie2007short, xiao2003fuzzy}. However, removing high-frequency components has led to discrepancies between the forecasts and ground truth data. To tackle this issue, Sun et al. \cite{sun2015novel} applied WT on the passenger flow dataset from the Beijing subway system and modeled both the details and smooth coefficients using SVR. While this approach improved short-term forecasts, it failed to capture the spatial dependencies in the dataset. To overcome this limitation, Zhang et al. introduced the Motif-GCRNN framework for generating spatiotemporal forecasts of traffic speed in Chengdu, China \cite{zhang2019wavelet}. This framework applies WT to the traffic speed data and generates the corresponding high-frequency and low-frequency components. They used a graph convolution recurrent neural network (GCRNN) to model the smooth components and the ARMA model for the detail coefficients. Despite its ability to capture both smooth and detailed fluctuations in the traffic speed dataset, Motif-GCRNN struggles with dynamic changes in spatiotemporal patterns. Additionally, the linear ARMA model used for the detail coefficients is insufficient for handling the nonlinearities often present in traffic data. To address these challenges, we propose the W-DSTAGNN approach, which integrates WT with a dynamic GNN capable of handling non-stationarity, nonlinearity, and dynamic spatiotemporal patterns in traffic data.

\section{Mathematical Preliminaries}\label{sec:maths}

\subsection{Wavelet Transformation}
Wavelet is a `small' wave-like oscillation, which is defined as a square-integrable function $\phi:\mathbb{R}\to\mathbb{R}$ such that $\int_{-\infty}^\infty \phi=0$ and $\int_{-\infty}^\infty \phi^2=1$. The second condition ensures that the wavelet is `localized' in time, thus allowing it to capture both the time and frequency of a signal,  unlike the Fourier transform, which is capable of capturing only the frequency of the signal. A wavelet transform converts a time series into a sequence of time-indexed observations, with each time series representing the original data in a particular frequency band. The wavelet transform can be done in two ways - continuous wavelet transform (CWT), which applies every possible wavelet to the original series, and discrete wavelet transform (DWT), which applies a finite number of wavelets at a specific time and location. In this study, we utilize the DWT approach that represents a series using an orthonormal basis and is widely used in hydrology \cite{hydrology}, epidemics \cite{epidemics}, geophysics \cite{wtransformer}, and economics \cite{sengupta2023forecasting}, among others. The DWT uses a dyadic grid. For scale parameter $i$ and shift parameter $k$, the equation for the decompositions using DWT is 
\begin{equation}
    c(i,k):=\sum_t U(t)(2^{\frac{i}{2}}\Psi(2^it-k))
\end{equation}  
where the sum is over the entire time series, $U$ is the original time series (or signal), and $\Psi$ is a mother wavelet.

\subsection{Maximal Overlap Discrete Wavelet Transform (MODWT)}

Application of DWT requires the sample size to be exactly a power of 2. Thus, a modified version of the DWT, namely maximal overlap discrete wavelet transform (MODWT), is adopted for decomposing arbitrary time series \cite{subtidal}. Both MODWT and DWT can accomplish multi-resolution analysis - a scale-based additive decomposition. However, in contrast to the usual DWT, in the MODWT, both wavelet and scaling coefficients are shift-invariant. Thus, circularly shifting the time series by any amount will circularly shift the MODWT details and smooth coefficients by a corresponding amount. This property is crucial, as it allows for the attention modules to be subjected to relatively `smoother' data, which makes it easier for them to capture the underlying pattern. Also, contrary to the DWT details and smooth, the MODWT details and smooth are associated with zero-phase filters, thus allowing the extraction of true signal from noise in a multiresolution analysis of the original time series. This allows for each of the attention blocks to have meaningful weights associated with them, thus leading to a robust framework \cite{subtidal}.

\begin{figure*}[htbp]
\centerline{\includegraphics[width=\linewidth]{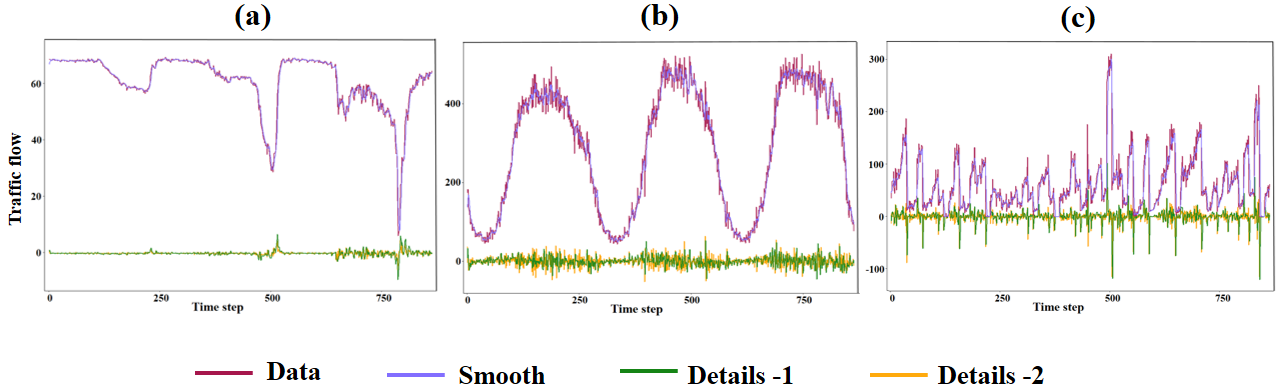}}
\caption{Traffic flow data (maroon) alongside its MODWT smooth (purple) and two details (green and orange) coefficients obtained at level 2 decomposition using a Haar wavelet filter. This data represents the traffic flow monitored by the third sensor of (a) PeMS-BAY, (b) PeMS03, and (c) PeMS04 dataset during the first three days of the training period.}
\label{fig:modwt_04}
\end{figure*}

To find the MODWT coefficients of level $j$ $(l=1,2,\ldots,J)$, the DWT coefficients are scaled and convolved with the original time series as follows.
\begin{equation}
    D_{j,t}=\sum_{l=0}^{L_j-1} \frac{d_{j,l}}{2^{j/2}} U_{(t-l)\text{mod}M'}, \
    S_{j,t}=\sum_{l=0}^{L_j-1} \frac{s_{j,l}}{2^{j/2}} U_{(t-l)\text{mod}M'}
\end{equation}
where $d_{j.l},s_{j,l}$ are the details and scaling coefficients of DWT and $L_j=(2^j-1)(L-1)+1$. Note that all the wavelet coefficients will have the same length as the original time series. Thus, the coefficients can be expressed in a matrix notation as 

\begin{equation}
    D_j = d_j U, \quad S_j= s_j U
\end{equation}
where $d_j$ and $s_j$ are square matrices of order $M'$ consists of the wavelet and scaling filters respectively. Hence, using MODWT, the original time series $U$ can be represented as

\begin{equation}
    U = \sum_{j = 1}^J d_j^T D_j + S_J^T S_J = \sum_{j = 1}^J \tilde{D}_j + \tilde{S}_J
\end{equation}
where $\tilde{D}_j; j = 1, 2, \ldots, J$ indicates the $j^{th}$ level high frequency details and $\tilde{S}_J$ represents the low frequency trend components. For a graphical illustration of the MODWT approach, we present the MODWT wavelet and scaling coefficients obtained by applying the transformation with the Haar filter at level 2 (J = 2) on selected sensor location from all three datasets in Figure \ref{fig:modwt_04}.

\begin{figure*}[htbp]
\centering
    \includegraphics[width=0.95\linewidth]{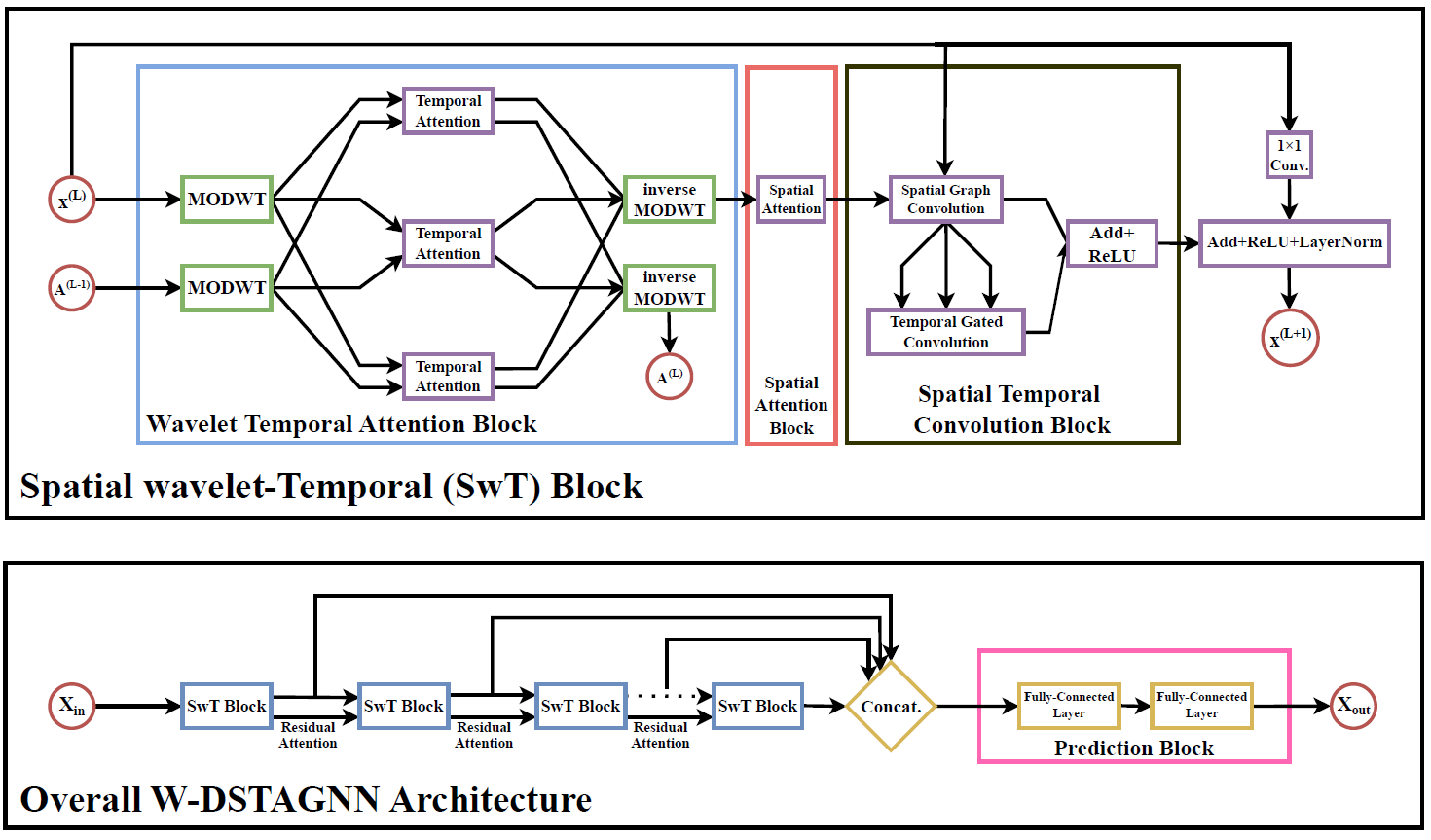}
    \caption{Detailed framework of the proposed W-DSTAGNN model. }
    \label{fig:architecture}
\end{figure*}

\section{Proposed Methodology: W-DSTAGNN}\label{sec:methodology}
The wavelet dynamic spatiotemporal aware graph neural network (W-DSTAGNN) architecture consists of stacked spatiotemporal attention blocks with a MODWT transformation as a pre-processing step in the temporal module and a prediction layer. To initialize the spatiotemporal blocks, we design the traffic road network in a graphical manner such that each sensor acts as a node in the graph, and the edges represent the corresponding connections between the nodes. In the W-DSTAGNN architecture, we compute the spatial association among the nodes using spatiotemporal aware distance (STAD), as proposed in \cite{dstagnn}. Thus the $(i, j)^{th}$ entry of the adjacency matrix ($A_{STAD}$) based on STAD can be represented as $A_{STAD}[i, j] = 1 - d_{STAD}(i, j)$ with $d_{STAD}(i, j)$ being the STAD between the corresponding sensors. To ensure the sparsity level in the adjacency matrix, we set the sparsity hyperparameter $P_{sp} = 0.01$ such that for each node $i$ ($i = 1, 2, \ldots, N$), the number of non-zero elements is $N_r = N \times P_{sp}$ which has the maximum value. Thus, the spatiotemporal relevance graph (STRG) created using these sparse connections has the adjacency matrix $\left[A_{STRG}\right]_{N \times N}$ with only $N_r \times N_r$ non-zero elements. Along with STAD, we utilize the wavelet-based spatiotemporal attention block to capture the dynamic characteristics of the spatial dependencies with changes in time.

In the wavelet temporal attention (wTA) block, we first preprocess the data using the MODWT-based multiresolution analysis. Then, we use multi-head self-attention layers to capture the long-range correlation in the time series data. This enhances the effectiveness of modeling the dynamic temporal dependencies between the nodes. Thus, we first apply the MODWT transformation to the input $X^{(l)} \in \mathbb{R}^{N \times c^{(l)} \times M}$ and the residual attention from the previous layer $A^{(l-1)}$ to generate the corresponding details ($\tilde{D}_j^{X^{(l)}}; j = 1, 2, \ldots, J$, $\tilde{D}_j^{A^{(l-1)}}; j = 1, 2, \ldots, J$) and smooth ($\tilde{S}_J^{X^{(l)}}$, $\tilde{S}_J^{A^{(l-1)}}$) coefficients. In the W-DSTAGNN approach, we aim to apply temporal attention to $X^{(l)},A^{(l-1)}$ by individually applying it to the details and smooth components and aggregating them using the inverse MODWT transformation. Thus, the wTA block can be mathematically represented as: 

\begin{align}
Y^{(l)} &= \textrm{IMODWT}\Big[ f\left(\tilde{D}_1^{X^{(l)}}, \tilde{D}_1^{A^{(l-1)}}\right),\ldots,\notag \\
& f\left(\tilde{D}_J^{X^{(l)}}, \tilde{D}_J^{A^{(l-1)}}\right),f\left(\tilde{S}_J^{X^{(l)}}, \tilde{S}_J^{A^{(l-1)}}\right)\Big]
\end{align}
where IMODWT is the Inverse MODWT, $f$ is the temporal attention applied to the details and smooth coefficients, which is defined as follows:

\begin{align}
    &f\left(\tilde{D}_j^{X^{(l)}}, \tilde{D}_j^{A^{(l-1)}}\right) = \notag \\
    &\textrm{LayerNorm}\left(\textrm{FullyConnected} \left([\mathcal{O}^1,\ldots,\mathcal{O}^H]+\tilde{D}_j^{X^{(l)}}\right)\right)
\end{align}
where $\mathcal{O}^h:=\textrm{softmax}(A^{(l)})V^{(l)}$, $\tilde{D}_j^{A^{(l)}}=\frac{Q^{(l)}K^{(l)^{T}}}{\sqrt{d_H}}+\tilde{D}_j^{A^{(l-1)}}$, $d_H=\frac{d}{H}$, $Q^{(l)}:=\tilde{D'}_j^{X^{(l)}}W_q^{(l)}, K^{(l)}:=\tilde{D'}_j^{X^{(l)}}W_k^{(l)}, V^{(l)}:=\tilde{D'}_j^{X^{(l)}}W_v^{(l)}$ and $\tilde{D'}_j^{X^{(l)}}\in\mathbb{R}^{c^{l-1}\times M\times N}$ is just a reshape of $\tilde{D}_j^{X^{(l)}}\in\mathbb{R}^{N\times c^{l-1}\times M}$, where $M$ is the number of time steps, $c^{l-1}$ is the feature dimension from the $(l-1)^{th}$ layer of the spatiotemporal block, and $N$ is the number of nodes  ($f(\tilde{S}_J^{X^{(l)}}, \tilde{S}_J^{A^{(l-1)}})$ is defined analogously). The spatial attention (SA) module receives $Y^{(l)}$ as the input and applies the self-attention mechanism to compute the dynamic spatial dependencies. Mathematically, the attention output $\mathcal{P}$ generated by the SA blocks can be represented as $\mathcal{P} = \textrm{SA}(Y^{(l)}) = [\mathcal{P}^1,\ldots,\mathcal{P}^H]$ with 

\begin{equation}
    \mathcal{P}^h=\textrm{softmax}\left(\frac{(Y_EW_k'^{(h)})(Y_EW_q'^{(h)})}{\sqrt{d_H}}+W_m^{(h)}\odot A_{STRG}\right)
\end{equation}
for $h = 1,2,\ldots,H$, where $Y_E=\textrm{Embedding}$ $(\textrm{Conv}(Y^{*}))$ and $Y^{*}\in\mathbb{R}^{c^{l-1}\times N\times M}$ is the transpose of $Y^{(l)}$ from the wTA layer. Thus, the output $\mathcal{P} = \left[\mathcal{P}^1, \mathcal{P}^2, \ldots, \mathcal{P}^H\right]$ denotes the spatiotemporal dynamic dependencies obtained by aggregating the output of the spatiotemporal attention modules.

The output of the wavelet spatiotemporal attention module is then passed into the spatial convolution block, a standard spatial graph convolution module that performs graph convolution based on Chebyshev polynomial approximation using the STAG. It is responsible for fully exploiting the traffic network's topological characteristics and learning the structure-aware node features.

\begin{equation}
    Z^{(l)} = g_\theta*_G X^{(l)} = g_\theta(L)X^{(l)} = \sum_{k=0}^K \theta_k(T_k(\Tilde{L})\odot P^{(k)})X^{(l)}
\end{equation}
where $\theta$ is learnable, $\Tilde{L}:=(2/\lambda_{max})(D-A_{STAG})-\mathbb{I}_N$, $D$ is the diagonal matrix with $D_{ii}=\sum_j A_{STAG}(i,j)$, $\lambda_{max}$ is the largest eigenvalue of $L=D-A_{STAG}$ and $T_k$ is the $k^{th}$ order Chebyshev polynomial. We finally process the output from the spatial layer using a temporal-gated convolutional network.  

We use the temporal gated convolution layer, which is composed of three Gated Tanh Units (GTU)  with different receptive fields. The forecast can be obtained as $X^{(l+1)} = \text{LayerNorm}(\text{ReLU}(X^{(l)} + Z_{out}))$ with

\begin{align}
    Z_{out} = &\textrm{ReLU}(\textrm{Concat}(\textrm{Pooling}(\Gamma_1 *_\tau Z^{(l)}),\notag \\
    &\textrm{Pooling}(\Gamma_2 *_\tau Z^{(l)}), \textrm{Pooling}(\Gamma_3 *_\tau Z^{(l)})) + Z^{(l)})
\end{align}
where $\Gamma_i$ is the convolution kernel of size $s_i$, $\Gamma *_\tau Q = \tanh(E)\odot\sigma(F)$ where $E, F$ are the first and second halves of $Q$ with respect to the channel dimension and concatenation and pooling is done such that $\frac{3M-(s_1+s_2+s_3-3)}{W}=M$. A pictorial illustration of the W-DSTAGNN architecture is given in Figure \ref{fig:architecture}.

\section{Experimental Setup}\label{sec:experiments}
In this section, we empirically evaluate the performance of the proposed W-DSTAGNN framework by conducting benchmark comparisons with state-of-the-art forecasters. The following subsections provide a brief description of the traffic forecasting datasets along with their statistical properties (Section \ref{sec:datasets}), the baseline models with their implementation strategies (Section \ref{sec:baselines}), key performance indicators (Section \ref{Sec:Performance}), experimental setup and benchmark comparisons (Section \ref{sec:comparison}), the statistical significance of the experimental results (Section \ref{Sec:MCB}), the influence of the hyperparameters (Section \ref{Sec:Hyperparameter_Tuning}), and uncertainty quantification of our proposal (Section \ref{Sec:Conformal_Predictions}).

\subsection{Datasets}\label{sec:datasets}
\begin{figure}[htbp]
    \centerline{\includegraphics[width=\linewidth]{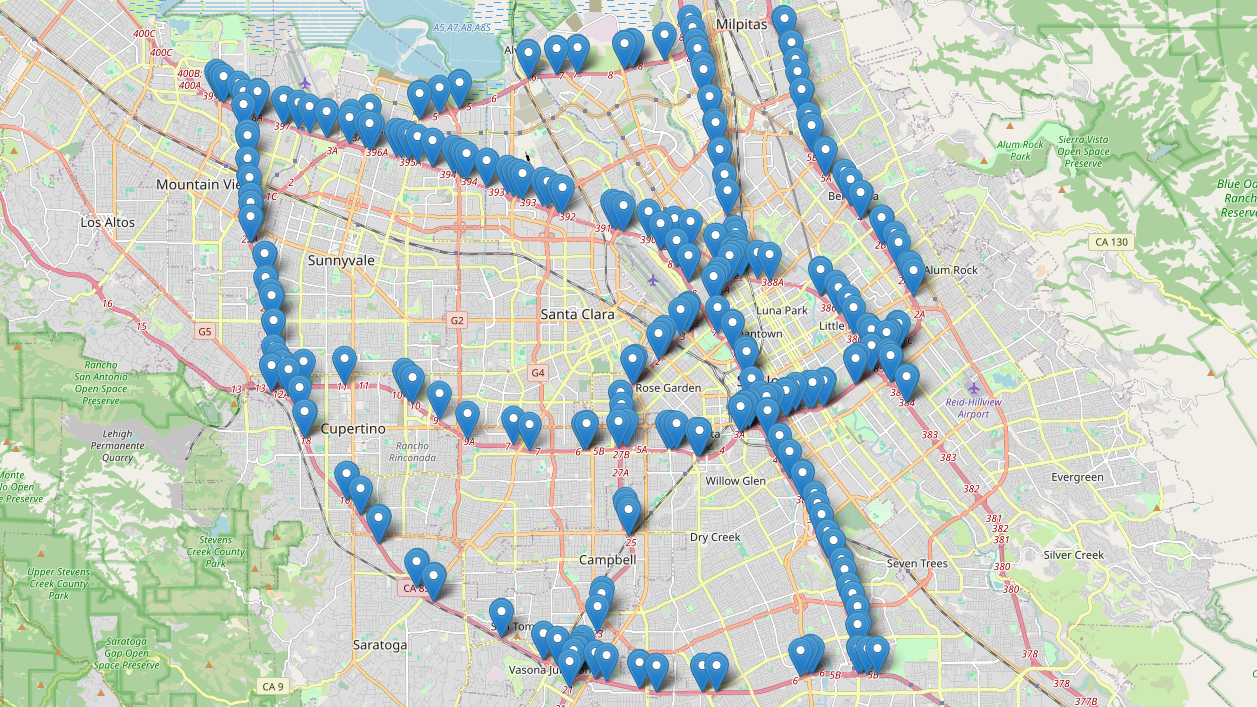}}
    \caption{Sensor distribution of PeMS-BAY dataset.}
    \label{fig:map}
\end{figure}

\begin{figure}
    \centering
    \includegraphics[width=\linewidth]{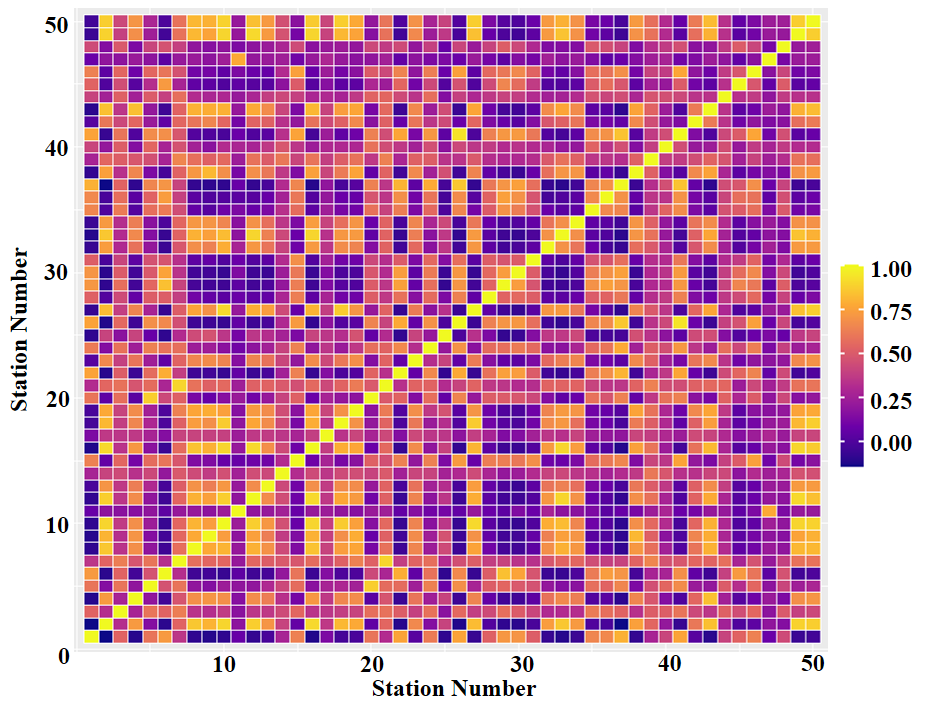}
    \caption{Heatmap of the correlation values for selected 50 sensor locations of PeMS-BAY dataset.}
    \label{fig:PeMS_BAY_Heatmap}
\end{figure}

To validate the performance of the W-DSTAGNN architecture, we conduct experiments on real-world traffic forecasting benchmark datasets acquired from the Caltrans PeMS. Our datasets include the PeMS-BAY dataset curated by \cite{dcrnn}, as well as the PeMS03 and PeMS04 datasets preprocessed by \cite{stsgcn}. All traffic datasets are gathered in real-time from several monitoring sensors strategically positioned throughout the freeway system across all major metropolitan areas of California, USA. The PeMS-BAY dataset accumulates information from 325 sensors in the Bay area, covering a six-month timespan from January 1, 2017, to May 31, 2017. The sensor distribution for the PeMS-BAY dataset is visualized in Figure \ref{fig:map}. For the PeMS03 dataset, 358 sensors were selected, and three months of data were collected from September 1, 2018, to November 30, 2018. For PeMS04, data from 307 selected sensors are collected for two months, spanning from January 1, 2018, to February 28, 2018. For all the datasets, aggregated traffic speed readings at 5-minute intervals are used in the experimental analysis. A summary of all datasets, including the number of sensors (nodes), number of samples, sample rate, and time range, is provided in Table \ref{tab:dataset_info}. Furthermore, we present a correlation heatmap of the traffic flow time series for selected sensor locations of the PeMS-BAY dataset in Figure \ref{fig:PeMS_BAY_Heatmap}. As depicted in the plot, most of the time, the series monitored by different sensors that possess significant correlations with their counterparts from other sensors. This highlights the presence of spatial and temporal interdependency in the dataset. For preprocessing the datasets, we apply normalization as \(X_{normalized} = \frac{X-mean(X)}{std(X)}\) to ensure zero mean and unit variance before training our forecasting models. In addition, we study several global features of these time series as listed below:

\textit{Stationarity} is a fundamental property of time series data, ensuring that its statistical characteristics, such as mean and variance, remain constant over time. This property is essential for maintaining the forecastability of the series and is a key assumption in various forecasting models. To assess the stationarity of the time series dataset, we use the Kwiatkowski–Phillips–Schmidt–Shin (KPSS) test, implemented via the `kpss.test' function from the \textit{tseries} package in R.

\textit{Linearity} is another important property of a time series for determining the appropriate forecasting model. A linear time series indicates that the data-generating process follows linear patterns. In this study, we apply Teraesvirta's neural network test to assess nonlinearity, using the `nonlinearityTest' function from the \textit{nonlinearTseries} package in R.

\textit{Long-term dependency} of a time series plays a significant role in probabilistic time series modeling. To determine the long-range dependency of the time series datasets, we compute the Hurst exponent using the `hurstexp' function of \textit{pracma} package in R.

\textit{Seasonality} of a time series indicates the repeating patterns of the series at regular intervals. To detect these recurring fluctuations in the dataset, we implement Ollech and Webel's combined seasonality test using the `isSeasonal' function from the \textit{seastests} package in R.

\textit{Normality} assumption of the observations plays a crucial role in the methodological development of statistical models in time series analysis. In our study, we perform the Anderson-Darling normality test using the `ad.test' function from the \textit{nortest} package in R to detect any departure of the time series observations from the normality assumption.

On performing the above-mentioned statistical tests at 5\% level of significance, we compute the global characteristics of the datasets and report the values in Table \ref{tab:dataset_info}. As the table highlights, all the time series observations from different sensor locations are long-term dependent and non-normal. Most time series from different datasets are non-linear and have seasonal patterns. Additionally, some of the series demonstrate non-stationary behavior. 

% The critical values of these statistical tests for a selected sensor are presented in Table \ref{tab:adf,kpss_results}. Overall, the test results depict that, according to the ADF test, all the time series corresponding to each sensor location for the PeMS-BAY, PeMS03, and PeMS04 datasets are stationary. However, according to the KPSS test, the time series corresponding to 246 sensors out of 325 for PeMS-BAY, 332 sensors out of 358 for PeMS03, and 298 sensors out of 307 for PeMS04 are stationary.

\begin{table*}[htbp]
    \centering
    \caption{Summary of the datasets with statistical features (values indicate the number of time series from different nodes that exhibit the statistical characteristic)}
    \label{tab:dataset_info}
    \setlength{\tabcolsep}{4.5pt} % Default value: 6pt, horizontal
    \begin{tabular}{|c|c|c|c|c|c|c|c|c|c|} %features : 3,1,3
        \hline
        Dataset & Nodes & Observations & Granularity & Time span & Stationarity & Linearity & Long-term dependency & Seasonality & Non-Normal\\
        \hline
        PeMS-BAY & 325 & 52116 & 5 min & 01/01/17-31/05/17 & 50 & 3 & 325 & 264 & 325\\
        PeMS03 & 358 & 26209 & 5 min & 01/09/18-30/11/18 & 287 & 0 & 358 & 358 & 358\\
        PeMS04 & 307 & 16992 & 5 min & 01/01/18-28/02/18 & 270 & 0 & 307 & 307 & 307\\
        \hline
    \end{tabular}
\end{table*}

% \begin{table}[htbp]
%     \centering
%     \caption{ADF (Critical Value at $1\%$ is -3.43) and KPSS (Critical Value at $10\%$ is 0.35) test results for sensor number 38.}
%     \label{tab:adf,kpss_results}
%     \setlength{\tabcolsep}{8pt} % Default value: 6pt, horizontal
%     \begin{tabular}{|c|c|c|}
%         \hline
%         Dataset & ADF Test Statistic & KPSS Test Statistic \\
%         \hline
%         PeMS-BAY & -30.56 & 0.92 \\
%         PeMS03 & -15.81 & 0.07 \\
%         PeMS04 & -16.05 & 0.12 \\
%         \hline
%     \end{tabular}
% \end{table}

\subsection{Baseline models}\label{sec:baselines}
In this section, we briefly explain the baseline models used in our experimental analysis and discuss their implementation strategies as adopted from \cite{dcrnn}. 
\begin{enumerate}
    \item \textbf{Autoregressive Integrated Moving Average} (ARIMA) is a classical time series forecasting approach used for tracking the linear trajectories in a time series data \cite{arima}. This framework applies differencing to obtain stationarity and models the lagged values of the original time series with the lagged error observations to generate forecasts. Implementation of the ARIMA model (with Kalman filter) is done with three lagged values and one lagged error using the \textit{statsmodel} \cite{statsmodel} python package.
    
    \item \textbf{Support Vector Regression} (SVR) is a supervised machine learning technique that fits an optimal hyperplane to forecast the time series observations \cite{svr}. To fit the SVR model, we transform the dataset from each node into a supervised setup, such that the future observations of the series relate to its previous 5 observations. Based on this transformed dataset, we fit the radial basis kernel-based SVR model with the loss penalty term $C = 0.1$ and generate the multi-step ahead forecasts using the \textit{sktime} python package.
    
    \item \textbf{Vector Autoregressive} (VAR) model is a linear multivariate forecasting technique that can model the pairwise relationships among different time series \cite{var}. This framework is a generalization of the univariate ARIMA model with the capability of incorporating feedback relationships from other variables. This framework treats each time series as an endogenous variable and utilizes auxiliary information from other time series to generate the corresponding forecasts. Implementation is done by setting the number of lags to 3, using the \textit{statsmodel} \cite{statsmodel} python package.
    
    \item \textbf{FC-LSTM} \cite{fclstm} is an encoder-decoder framework using LSTM with peephole. There are two recurrent layers in the encoder and the decoder. In each recurrent layer, there are 256 LSTM units. The L1 weight decay is $2e^{-5}$ and L2 weight decay is $5e^{-4}$. The batch size is 64, and the loss function is MAE. The initial learning rate is $1e^{-4}$, which reduces by $1/10$ every 10 epochs starting from the 20$^{th}$ epoch. Early stop is also performed by looking at the validation error.
    
    \item \textbf{Diffusion Convolutional Recurrent Neural Network} (DCRNN) model is a spatiotemporal forecasting technique that integrates bidirectional random walks on the graphs with recurrent neural networks \cite{dcrnn}. This architecture performs diffusion convolution on the graphs to capture the spatial dependencies and model them using an encoder-decoder architecture with scheduled sampling techniques to generate long-term spatiotemporal forecasts. In our study, we adopt the implementation of the DCRNN approach from the open-access \href{https://github.com/liyaguang/DCRNN}{GitHub repository} of \cite{dcrnn}.

    \item \textbf{Spatial-Temporal Graph Convolutional Networks} (STGCN) combine graph convolution with gated temporal convolutions to effectively capture both spatial and temporal dependencies in spatiotemporal datasets \cite{stgcn}. The model consists of multiple convolutional layers, allowing faster training and fewer parameters. The implementation of the STGCN model is based on the \href{https://github.com/VeritasYin/STGCN_IJCAI-18}{GitHub repository} provided by \cite{stgcn}.
     
    \item \textbf{Spatial-Temporal Synchronous Graph Convolutional Networks} (STSGCN) is a robust spatiotemporal modeling technique that captures the localized spatiotemporal correlations along with the heterogeneities in the dataset \cite{stsgcn}. This framework applies several graph convolution operations to model the spatial dependencies and utilizes two fully connected layers for the temporal patterns. We adopt the implementation from the \href{https://github.com/Davidham3/STSGCN}{GitHub repository} of \cite{stsgcn} to apply the STSGCN model.
    
    \item \textbf{GraphWavenet} (GWN), introduced in \cite{graphwavenet}, integrates graph convolution with dilated casual convolution to understand spatiotemporal dependencies. This model develops an adaptive dependency matrix through node embeddings to capture the hidden spatiotemporal patterns in the dataset efficiently. This framework employs a stacked 1D convolution layer with many receptive fields to model long-range dependencies in the temporal window. To implement this model, we adopted the code available at the \href{https://github.com/nnzhan/Graph-WaveNet}{GitHub repository} of \cite{graphwavenet}.
    
    \item \textbf{Adaptive Graph Convolutional Recurrent Network} (AGCRN) framework performs spatiotemporal forecasting by integrating three key components: a Node Adaptive Parameter Learning (NAPL) module, a Data-Adaptive Graph Generation (DAGG) module, and a recurrent network \cite{agcrn}. The NAPL and DAGG modules are designed to capture node-specific patterns and the interdependencies between the traffic series. At the same time, the recurrent network focuses on modeling the temporal dynamics within the dataset. The implementation of the AGCRN model is based on the \href{https://github.com/LeiBAI/AGCRN}{GitHub repository} provided by \cite{agcrn}.
    
    \item \textbf{Graph Multi-Attention Network} (GMAN) employs an encoder-decoder architecture with a transformer-based attention mechanism for spatiotemporal forecasting \cite{gman}. The encoder consists of multiple spatiotemporal blocks that process the input, which is then transformed using the attention mechanism. The spatiotemporal attention blocks within the decoder generate the forecasts. The implementation of the GMAN model is based on the \href{https://github.com/zhengchuanpan/GMAN}{GitHub repository} provided by \cite{gman}.
    
    \item \textbf{Dynamic Spatial-Temporal Aware Graph Neural Network} (DSTAGNN) introduces a data-driven approach to capture complex dynamic spatiotemporal dependencies in road networks \cite{dstagnn}. This modified architecture leverages a multi-head attention mechanism to capture dynamic spatial relationships among nodes, while multi-scale gated convolutions are used to model dynamic temporal patterns. The implementation of DSTAGNN is based on the \href{https://github.com/SYLan2019/DSTAGNN}{GitHub repository} of \cite{dstagnn}.
\end{enumerate}

\subsection{Performance measure}\label{Sec:Performance}
To measure the performance of different forecasting frameworks, we use three performance indicators, namely, mean absolute error (MAE), mean absolute percentage error (MAPE), and root mean squared error (RMSE). These metrics can be computed as: 
{\[\text{MAE}(y,\hat{y})= \frac{1}{n}\sum_{i=1}^n |y_i-\hat{y}_i|, \; \]
\[\text{MAPE}(y,\hat{y})=\frac{1}{n}\sum_{i=1}^n \frac{|y_i-\hat{y}_i|}{y_i}, \;\]
\[\text{RMSE}(y,\hat{y})=\sqrt{\frac{1}{n}\sum_{i=1}^n (y_i-\hat{y}_i)^2},\]}
%\[\Scale[0.8]{MAE(x,\hat{x})= \frac{1}{n}\sum_{i=1}^n |x_i-\hat{x}_i|, \; MAPE(x,\hat{x})=\frac{1}{n}\sum_{i=1}^n \frac{|x_i-\hat{x}_i|}{x_i},\; RMSE(x,\hat{x})=\sqrt{\frac{1}{n}\sum_{i=1}^n (x_i-\hat{x}_i)^2}}\]
%$${ MAE(x,\hat{x})= \frac{1}{n}\sum_{i=1}^n |x_i-\hat{x}_i|, \; MAPE(x,\hat{x})=\frac{1}{n}\sum_{i=1}^n \frac{|x_i-\hat{x}_i|}{x_i},}$$ 
%$${ RMSE(x,\hat{x})=\sqrt{\frac{1}{n}\sum_{i=1}^n (x_i-\hat{x}_i)^2}}$$  
\noindent where $y=(y_1,\ldots,y_n)$ is the testing data (ground truth), and $\hat{y}=(\hat{y}_1,\ldots,\hat{y}_n)$ is the corresponding forecast. By general convention, the model with the least performance measure is the `best' forecasting model. We reported the testing errors computed between the test data and the data forecasted by the model. %If $x=(x_1,\ldots,x_n)$ is the testing data (ground truth), and $\hat{x}=(\hat{x}_1,\ldots,\hat{x}_n)$ is the predicted data, then the metrics stated above are defined as follows.
%\[ MAE(x,\hat{x}):= \frac{1}{n}\sum_{i=1}^n |x_i-\hat{x}_i| \qquad
% MAPE(x,\hat{x}):=\frac{1}{n}\sum_{i=1}^n \frac{|x_i-\hat{x}_i|}{x_i} \] 
%\[ RMSE(x,\hat{x}):=\sqrt{\frac{1}{n}\sum_{i=1}^n (x_i-\hat{x}_i)^2} \]

\subsection{Experimental setup and performance comparison}\label{sec:comparison}
In the experimental setup, to ensure a fair comparison with the baseline models, we apply a train-validation-test split to our spatiotemporal datasets. The PeMS-BAY dataset is split in a 7:1:2 ratio, while the PeMS03 and PeMS04 datasets are divided using a 6:2:2 ratio. For forecasting, we use one hour of historical data to predict traffic flow for the following hour. The training of the W-DSTAGNN and baseline models on the spatiotemporal datasets is performed using the T4 GPU on Google Colab Pro+. Based on the validation loss, we tune the hyperparameters in the W-DSTAGNN framework. In our experiments, the order of Chebyshev's polynomial ($k$) is set to 3, indicating that the spatial attention layer uses 3 attention heads. The temporal gated convolution layer employs 32 convolution kernels of sizes $\{s_1, s_2, s_3\}$ = $\{3, 5, 7\}$, along with a pooling layer of window size 2. Additionally, the spatial graph convolution layer utilizes 32 convolution kernels. In the wavelet temporal attention layer, we apply wavelet decomposition of level 2 and use 3 attention heads. The model architecture consists of 4 stacked spatiotemporal blocks, each containing a spatiotemporal attention module with 32 attention heads. For training, we utilize the Huber loss function with the Adam optimizer. The model is trained for 100 epochs with a learning rate of $10^{-4}$ and a batch size of 32. For certain baseline models, we use the forecasts reported in the seminal works of \cite{stsgcn, dcrnn, dstagnn}. 

\begin{figure}[htbp]
    \centerline{\includegraphics[width=1\linewidth]{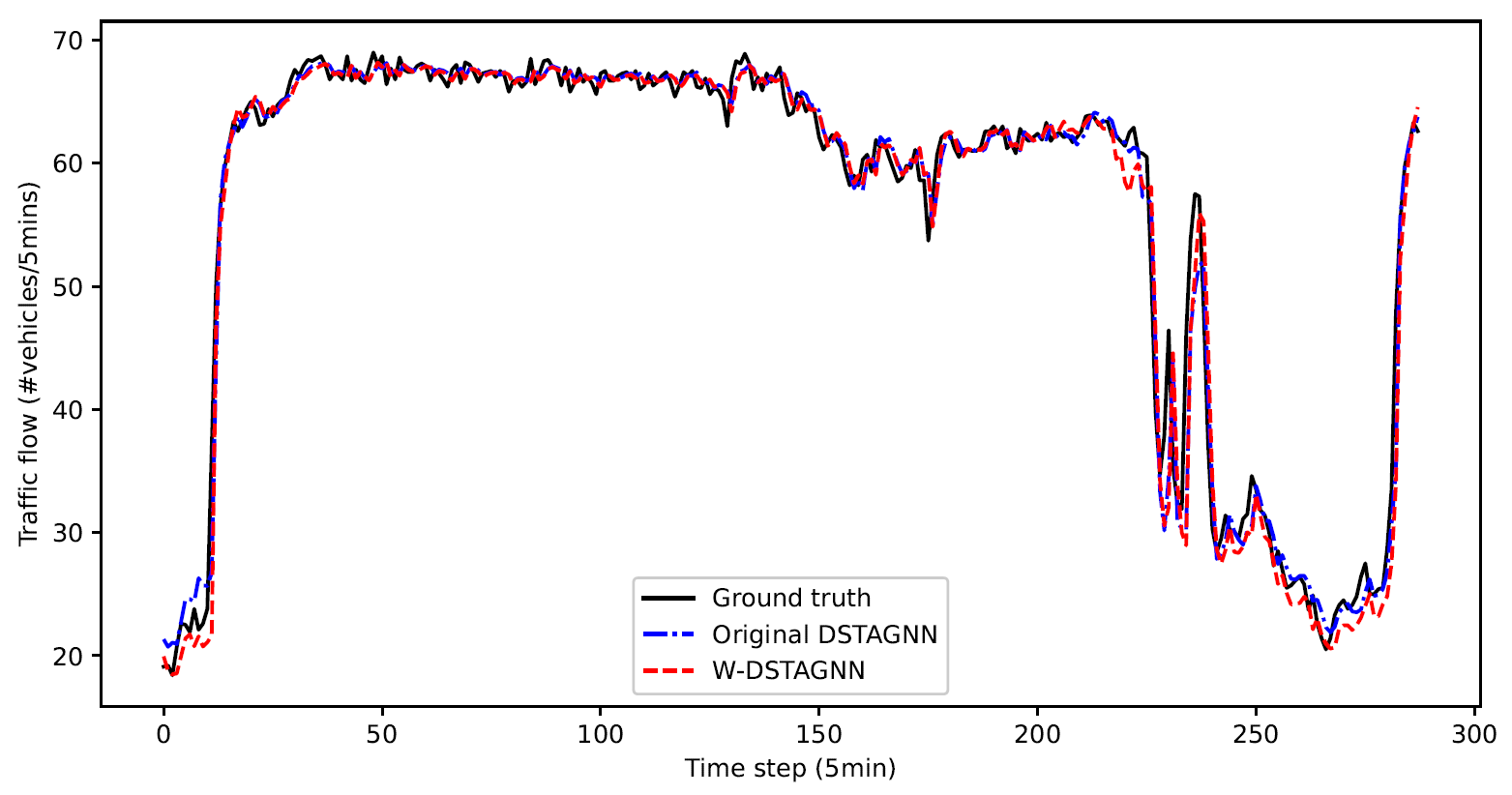}}
    \caption{Ground truth data for node 17 of PeMS-BAY on 1st testing day (black) and their corresponding forecasts generated by W-DSTAGNN (red) and DSTAGNN (blue).}
    %Prediction Curve (1st testing day, node 17 of PeMS-BAY)}
    \label{fig:predcurve_bay}
\end{figure}

\begin{figure}[htbp]
    \centerline{\includegraphics[width=1\linewidth]{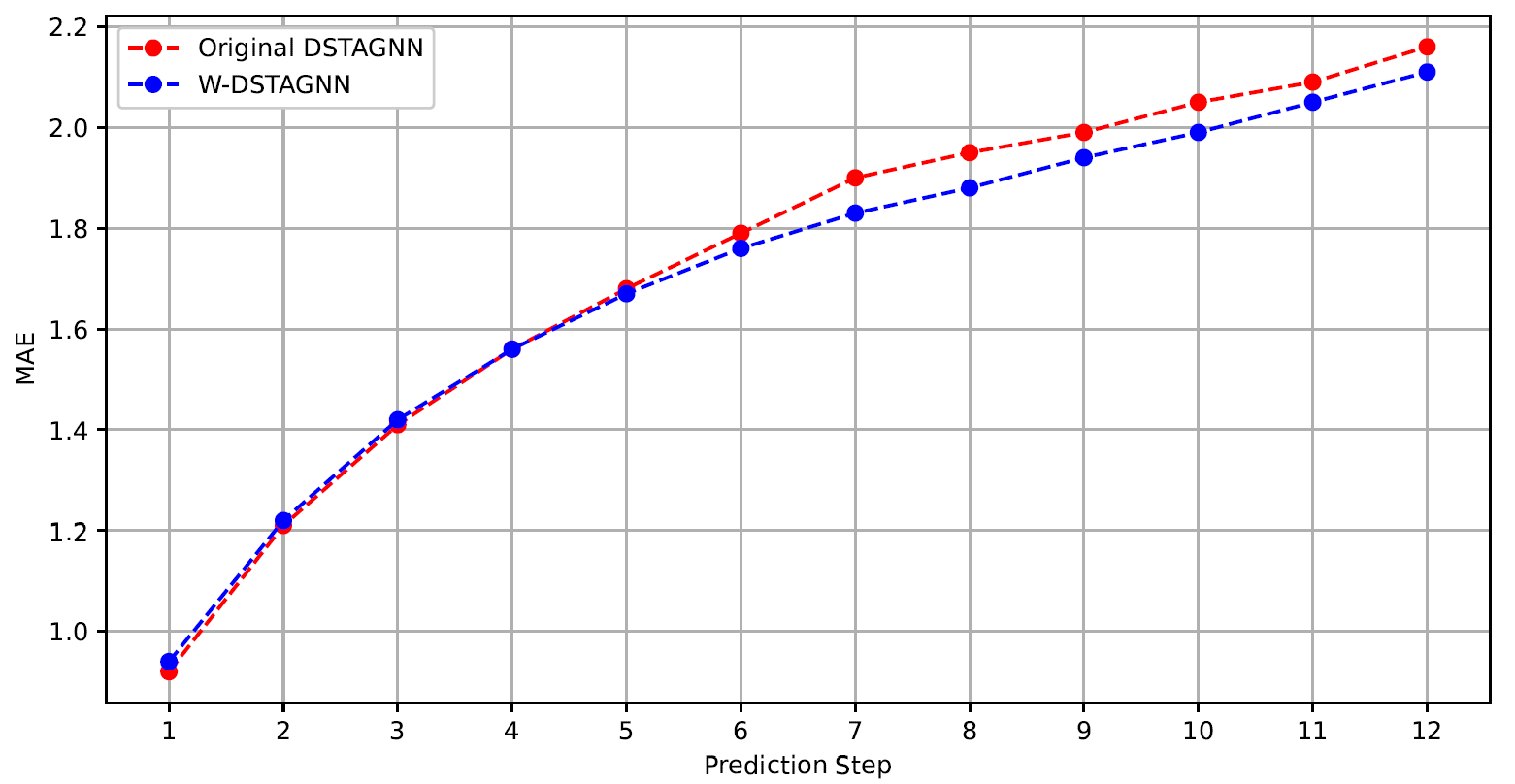}}
    \caption{Step-wise forecast error (MAE) comparison between DSTAGNN (red) and W-DSTAGNN (blue) for the PeMS-BAY dataset.}
    \label{fig:mae_pps_bay}
\end{figure}

Table \ref{tab:exp_results} presents the performance comparison of the proposed W-DSTAGNN architecture with various baseline methods for forecasting 1-hour (12 steps) ahead traffic conditions across different locations. The results highlight that W-DSTAGNN consistently provides more accurate forecasts for the traffic flow datasets based on different key performance metrics. Specifically, for the PeMS-BAY dataset, W-DSTAGNN outperforms other models with over 96\% forecast accuracy in terms of the MAPE metric. Similarly, the RMSE and the MAE metrics demonstrate the superiority of our model. In the PeMS03 dataset, W-DSTAGNN achieves the best forecasting performance across all three accuracy measures, highlighting its robustness. For the PeMS04 dataset, the DSTAGNN model shows competitive results with W-DSTAGNN based on MAPE values and GMAN slightly surpasses W-DSTAGNN with a margin of 1.34\% for the MAE metric. Nevertheless, W-DSTAGNN still achieves the highest accuracy for the RMSE metric. Moreover, from the above experimental results, it is evident that the performance of the conventional architectures like ARIMA, SVR, VAR, and FC-LSTM drastically drops in comparison to the proposed W-DSTAGNN approach. This is due to their ability to capture only the temporal correlations, ignoring the spatial dependencies. However, for the other spatiotemporal models, their better accuracy measures over the temporal architectures highlight the importance of modeling spatiotemporal dependencies. Moreover, to emphasize the importance of wavelet transformation in the W-DSTAGNN model, we compare the performance improvement of W-DSTAGNN over standard DSTAGNN by computing
\[\frac{metric(\text{DSTAGNN})-metric(\text{W-DSTAGNN})}{metric(\text{DSTAGNN})}\times 100\%.\]
The performance enhancement values as reported in table \ref{tab:exp_results} indicate that the W-DSTAGNN framework improves the RMSE score by a maximum of 2.51\%, MAPE by 1.53\%, and MAE by 1.67\% of DSTAGNN. This improvement in the performance measures is primarily attributed to the use of MODWT decomposition in the DSTAGNN architecture, which helps segregate signals from noise in the input data. A higher node count allows for redundancies in the form of relations between the nodes, which do not get removed by MODWT. Thus, the temporal attention block is allowed to look for patterns in the transformed data, while the spatial attention is free to look into the intricate inter-node patterns that were missed by temporal attention, leading to better learning of the overall pattern in the dataset. A graphical illustration of the ground truth, along with the forecasts generated by the DSTAGNN and W-DSTAGNN for the first testing day of sensor 17 of PEMS-BAY, are presented in Figure \ref{fig:predcurve_bay}. Additionally, we present the MAE values of DSTAGNN and W-DSTAGNN models in each 5-minute interval for a 1-hour forecast period in Figure \ref{fig:mae_pps_bay}.

\begin{table*}[htbp]
    %DSTAGNN should have a star, as that is the model we reimplemented
    \centering
    \caption{Experiment Results show that the proposed W-DSTAGNN model \textbf{outperforms} all baseline models. \\ (* denotes reimplementation)}
    \label{tab:exp_results}
    \renewcommand{\tabcolsep}{9pt}
    \begin{tabular}{|l|ccc|ccc|ccc|}%ccc|ccc|}
    \hline
      \multicolumn{1}{|c}{Baselines} &
      \multicolumn{3}{|c}{PeMS-BAY} &
      \multicolumn{3}{|c}{PeMS03} &
      \multicolumn{3}{|c|}{PeMS04} \\
      %\multicolumn{3}{|c}{PeMS07} &
      %\multicolumn{3}{|c|}{PeMS08} \\
      %\midrule
      & MAE & MAPE(\%) & RMSE & MAE & MAPE(\%) & RMSE & MAE & MAPE(\%) & RMSE\\
      %& MAE & MAPE(\%) & RMSE & MAE & MAPE(\%) & RMSE\\
      \hline
    ARIMA~\cite{arima} & 3.38 & 8.30 & 6.50 & 35.31 & 33.78 & 47.59 & 33.73 & 24.18 & 48.80 \\ % & 38.17	&	19.46	&	59.27	&	31.09	&	22.73	&	44.32 \\
    SVR~\cite{svr} & 3.28 & 8.00 & 7.08 & 21.97 & 21.51 & 35.29 & 28.70 & 19.20 & 44.56 \\ % &	32.49	&	14.26	&	50.22	&	23.25	&	14.64	&	36.16 \\
    VAR~\cite{var} & 2.93 & 6.50 & 5.44 & 23.65 & 24.51 & 38.26 & 23.75 & 18.09 & 36.66 \\ % &	75.63	&	32.22	&	115.24	&	23.46	&	15.42	&	36.33\\
    FC-LSTM~\cite{fclstm} & 2.37 & 5.70 & 4.96 & 21.33 & 22.33 & 35.11 & 26.24 & 19.30 & 40.49 \\ % &	29.96	&	14.34	&	43.94	&	22.2	&	15.02	&	33.06 \\
    DCRNN~\cite{dcrnn} & 2.07 & 4.90 & 4.74 & 18.18 & 18.91 & 30.31 & 24.70 & 17.12 & 38.12 \\ % &	25.3	&	11.66	&	38.58	&	17.86	&	11.45	&	27.83\\
    STGCN~\cite{stgcn} & 2.49 & 5.79 & 5.69 & 17.49 & 17.15 & 30.12 & 22.70 & 14.59 & 35.55 \\ % &	25.38	&	11.08	&	38.78	&	18.02	&	11.40	&	27.83\\
    STSGCN~\cite{stsgcn} & 2.11 & 4.96 & 4.85 & 17.48 & 16.78 & 29.21 & 21.19 & 13.90 & 33.65 \\ % &	24.26	&	10.21	&	39.03	&	17.13	&	10.96	&	26.80 \\
    GWN~\cite{graphwavenet} & 1.95 & 4.63 & 4.52 & 19.85 & 19.31 & 32.94 & 25.45 & 17.29 & 39.70 \\ % &	26.85	&	12.12	&	42.78	&	19.13	&	12.68	&	31.05 \\
    AGCRN~\cite{agcrn} & 1.96 & 4.64 & 4.54 & 15.98 & 15.23 & 28.25 & 19.83 & 12.97 & 32.26 \\ %&	\textbf{21.29}	&	\textbf{8.97}	&	35.12	&	15.95	&	10.09	&	25.22\\
    GMAN~\cite{gman} & 1.86 & 4.31 & 4.32 & 16.87 & 18.23 & 27.92 & \textbf{19.14} & 13.19 & 31.60 \\ %&	21.97	&	9.05	&	35.10	&	\textbf{15.31}	&	10.13	&	24.92\\
    % STAEformer & 1.88 & 4.41 & 4.34 & 15.35 & 15.18 & 27.55 & 18.22 & 11.98 & 30.18 \\
    DSTAGNN~\cite{dstagnn} & 1.72* & 3.92* & {3.98}* & 15.57 & 14.68 & 27.21 & 19.30 & \textbf{12.70} & 31.46 \\ %&	21.42	&	9.01	&	\textbf{34.51}	&	15.67	&	\textbf{9.94}	&	\textbf{24.77}\\
    % MTGNN & 1.94 & 4.53 & 4.49 & 14.85 & 14.55 & 25.23 & 19.17 & 13.37 & 31.70 \\
    % AutoSTS & 1.83 & 4.41 & 4.27 & 14.61 & 14.18 & 24.71 & 18.76 & 12.84 & 30.31 \\
    % \hline
    \textbf{W-DSTAGNN} (Proposed) & \textbf{1.70} & \textbf{3.86} & \textbf{3.88} & \textbf{15.31} & \textbf{14.49} & \textbf{26.59} & 19.30 & \textbf{12.70} & \textbf{31.28} \\ %&	21.8	&	9.37	&	35.02	&	15.88	&	10.06	&	24.91\\ \hline
    Performance Improvement & 1.16\% & 1.53\% & 2.51\% & 1.67 \% & 1.29\% & 2.28\% & 0.00\% & 0.00\% & 0.57\% \\ \hline
  \end{tabular}
\end{table*}

% \begin{table}[htbp]
%     \centering
%     \caption{Improvement(\%) of W-DSTAGNN over DSTAGNN.}
%     \label{tab:improvement}
%     \setlength{\tabcolsep}{8pt} % Default value: 6pt, horizontal
%     \begin{tabular}{|c|c|c|c|c|}
%         \hline
%         Dataset & MAE & MAPE(\%) & RMSE & Mean \\
%         \hline
%         PeMS-BAY & 1.16 & 1.53 & 0 & 0.90\\ %79/325
%         PeMS03 & 1.67 & 1.29 & 2.28 & 1.75\\ %26/358
%         PeMS04 & -0.52 & 0 & 0.57 & 0.02\\ %9/307
%         \hline
%     \end{tabular}
% \end{table}

\subsection{Statistical Significance}\label{Sec:MCB}
\begin{figure*}[htbp]
    \centerline{\includegraphics[width=0.85\linewidth]{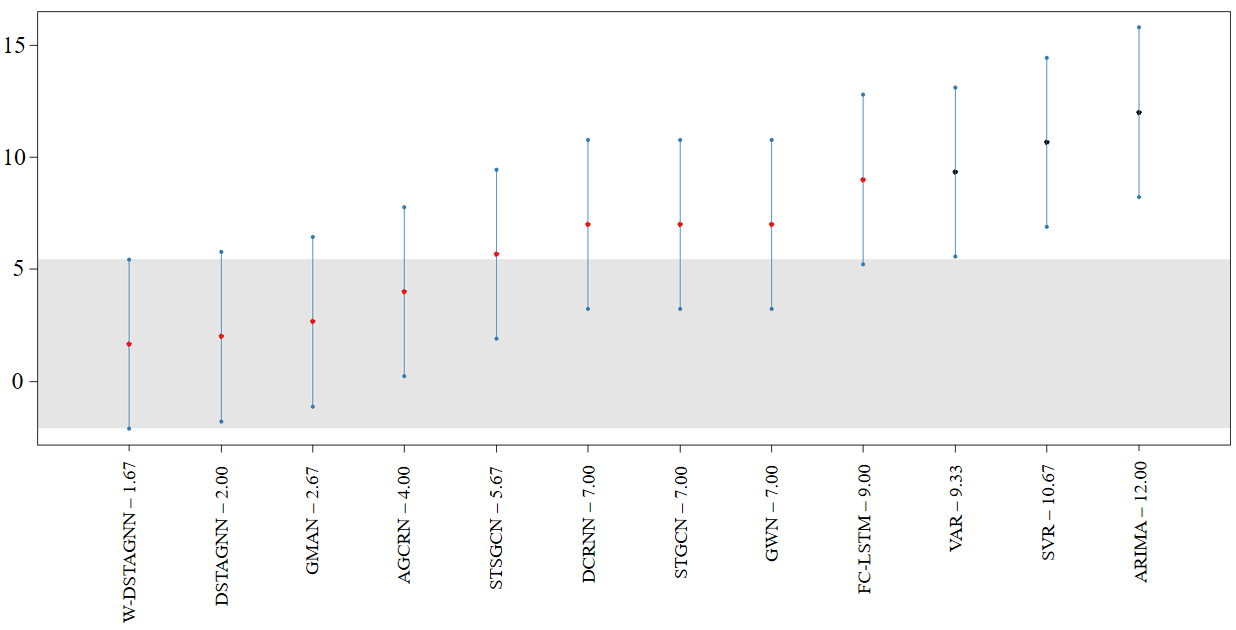}}
    \caption{Multiple comparisons with the best analysis for the three benchmark datasets in terms of MAE metric. In the plot, W-DSTAGNN - 1.67 indicates that the average rank of W-DSTAGNN is 1.67, similar to others.}
    \label{fig:mcb_mae}
\end{figure*}

Furthermore, to validate the statistical significance of our experimental evaluations, we used multiple comparisons with the best (MCB) test \cite{mcb}. This non-parametric test ranks the models based on their relative performance in terms of a specific metric and identifies the model with the minimum rank as the `best' performing approach. In the subsequent step, it determines the critical distance for each of the $\mathcal{M}$ competing forecasters as $\Xi_{\gamma} \sqrt{\frac{\mathcal{M}\left(\mathcal{M}+1\right)}{6 \mathcal{D}}}$ where $\mathcal{D}$ represents the number of datasets and $\Xi_{\gamma}$ is the critical value of the Tukey distribution at level $\gamma$. This distribution-free test treats the critical distance of the `best' performing model as the reference value of the test and compares the performance of the other models with this value. Figure \ref{fig:mcb_mae} presents the MCB test result computed based on the MAE metric. This plot highlights that our proposal is the `best' performing model as it achieves the minimum average rank of 1.67, followed by other spatiotemporal architectures. Moreover, the critical distance of the W-DSTAGNN model (shaded region) represents the reference value of the test. Since the critical distance of most of the temporal forecasting models lies well above the reference value, we can conclude that their performance is significantly inferior to the W-DSTAGNN model.

\subsection{Effect of Hyperparameters}\label{Sec:Hyperparameter_Tuning}
\begin{figure}[htbp]
    \centerline{\includegraphics[width=\linewidth]{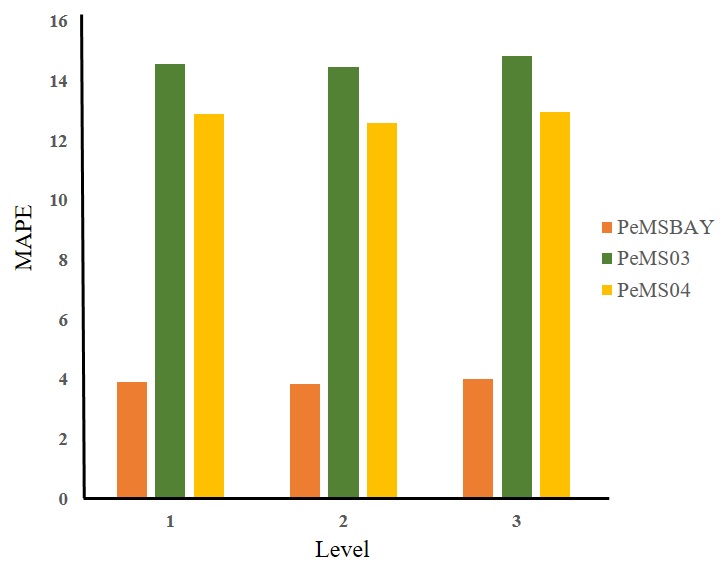}} 
    \caption{Forecast performance (MAPE) of W-DSTAGNN for PeMS-BAY (orange), PeMS03 (green), and PeMS04 (yellow) datasets with varying the level of MODWT decomposition in the wavelet temporal attention block.}
    \label{fig:hyperparameter}
\end{figure}

To ensure a fair comparison between the proposed W-DSTAGNN and the baseline DSTAGNN approaches, we use the same hyperparameters (as discussed in Section \ref{sec:comparison}) for both models. Additionally, we investigate the impact of different MODWT decomposition levels on the forecast performance of the W-DSTAGNN architecture. Increasing the MODWT level enhances the number of temporal attention blocks, consequently increasing the training time. To optimize forecast performance with computational complexity, we limit the decomposition to 3 levels. Figure \ref{fig:hyperparameter} illustrates the forecast performance of W-DSTAGNN using the MAPE metric for various MODWT levels. As evident from the plot, the second-level decomposition, with one low-frequency and two high-frequency components, achieves the best performance across all traffic datasets. This result suggests that the true signals in traffic flow data are effectively captured with a smooth series and two detail series, where the first detail series represents the most rapid fluctuations, the second captures moderate variations, and the remaining fluctuations can be treated as noise.

% \begin{table}[htbp]
%     \centering
%     \caption{Hyperparameter tuning results. It can be seen that level \textbf{2} gives the \textbf{best} results.}
%     \label{tab:hyperparameter}
%     \renewcommand{\tabcolsep}{2pt}
%     \begin{tabular}{|c|ccc|ccc|ccc|}
%     \hline
%       \multicolumn{1}{|c}{Level} &
%       \multicolumn{3}{|c}{PeMS-BAY} &
%       \multicolumn{3}{|c}{PeMS03} &
%       \multicolumn{3}{|c|}{PeMS04} \\
%       %\midrule
%       & MAE & MAPE & RMSE & MAE & MAPE & RMSE & MAE & MAPE & RMSE \\
%       \hline
%     1 & \textbf{1.70} & 3.88 & \textbf{3.87} & 15.53 & 14.51 & 26.96 & 19.45 & 12.89 & 31.33 \\
%     2 & \textbf{1.70} & \textbf{3.86} & 3.88 & \textbf{15.31} & \textbf{14.49} & \textbf{26.59} & \textbf{19.40} & \textbf{12.70} & \textbf{31.28} \\
%     3 & 1.72 & 4.01 & 3.96 & 15.62 & 14.83 & 27.29 & 19.56 & 12.96 & \textbf{31.28} \\
%     \hline
%   \end{tabular}
% \end{table}

\subsection{Conformal Predictions}\label{Sec:Conformal_Predictions}

\begin{figure*}[htbp]
    \centerline{\includegraphics[width=\linewidth]{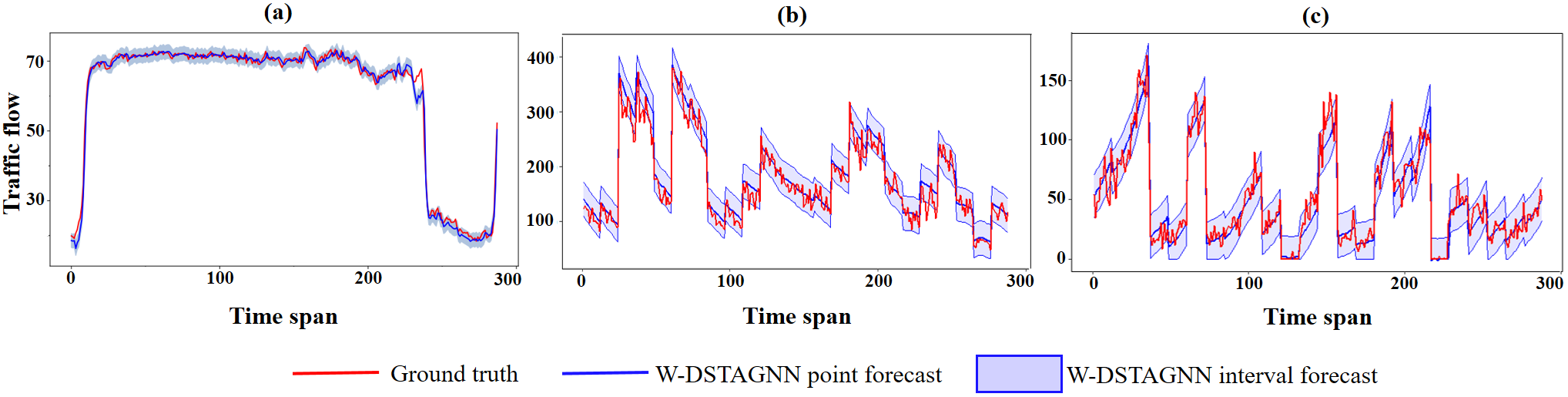}}
    \caption{Ground truth traffic dataset (red line) with the corresponding point forecasts (blue line), and 90\% conformal prediction interval (blue shaded region) generated by the W-DSTAGNN architecture for the first testing day of (a) PeMS-BAY (node 56), (b) PeMS03 (node 1), and (c) PeMS04 (node 1) datasets.}
    \label{fig:conformal_bay}
\end{figure*}

Alongside the point estimates, we utilize the conformal prediction approach to quantify the uncertainties associated with our proposal. The conformal prediction approach translates point estimates into prediction regions in a distribution-free, model-agnostic manner, guaranteeing convergence \cite{cp_book}. In the time series forecasting setup, this method leverages the sequential nature of the time series dataset. Given the input series $\{\mu_t\}_{t=1}^\mathcal{N}$ from a sensor, we fit the W-DSTAGNN and the uncertainty model $\xi$ on its lagged observations $\{\bar{\mu}_{t-1}\}$ to generate the scalar notion of uncertainty. Thus, the conformal score can be computed as follows:
\begin{equation*}
    \mathcal{W}_t = \frac{\left|\mu_t - \text{W-DSTAGNN}\left(\bar{\mu}_{t-1}\right)\right|}{\xi\left(\bar{\mu}_{t-1}\right)}.
\end{equation*}
Since $\mu_t$ possess a sequential pattern, thus we utilize a weighted conformal method with a fixed $\alpha$-sized window $\delta_t = \mathbbm{1}\left(t' \geq t - \alpha\right), \forall t' < t$ to compute the conformal quantile $\left(\text{CQ}_t\right)$ as 
\begin{equation*}
    \text{CQ}_t = \inf\left\{\mathcal{Q}: \frac{1}{\min\left(\alpha, t' -1\right) + 1} \sum_{t' = 1}^{t-1} \mathcal{W}_{t'} \delta_{t'} \geq 1- \beta \right\}.
\end{equation*}
Thus the $100*\left(1- \beta\right)\%$ conformal prediction interval based on these weighted quantiles is given by:
\begin{equation*}
    \text{W-DSTAGNN}\left(\bar{\mu}_{t-1}\right) \pm \text{CQ}_t \xi\left(\bar{\mu}_{t-1}\right).
\end{equation*}
In this study, we compute the conformal prediction interval with $90\%$ uncertainty quantification capacity for the first testing day of selected sensor locations of all three datasets and present it in Figure \ref{fig:conformal_bay}. To restrict data leakage and generate reliable prediction intervals, we calculate the residuals of the trained model that are applied to a calibration (validation) set.

\section{Limitations and future scope of this study}\label{sec:limitations}
As our proposed spatiotemporal forecasting approach has two key components namely wavelet decomposition and dynamic spatiotemporal GNN, therefore, there are a few limitations of the proposal. The proposed model's complexity is higher than the state-of-the-art DSTAGNN model which might obstruct the scalability of the proposal for very large datasets. However, for medium and small sample-sized datasets this problem will not arise. In addition to this, the improvement in terms of RMSE is around 2\%. This is because we performed all the experiments based on the benchmark papers where a 1-hour ahead (12 steps) forecast window is used for the performance evaluation. As our proposal is most suited for long-term forecasting, therefore, we may expect more significant improvement in terms of performance metrics for longer forecast horizons. This study promises several future scopes of research:
(a) Implementation of our method for long-range spatiotemporal forecasting of traffic or other domain-specific datasets;
(b) Incorporation of other causal variables that impact traffic flow inside the forecasting framework;
(c) Improvement of W-DSTAGNN using faster versions of spherical harmonic transformation that results in reduced computational complexity.

\section{Conclusion}\label{sec:conclusion}
In this paper, we presented a spatiotemporal deep learning model to perform traffic forecasting integrating wavelet decomposition with a temporal attention mechanism. Our ensemble approach outperformed other state-of-the-art models on several real-world traffic flow datasets, specifying their potential to tour spatiotemporal structures from the input time series. The key advantage of our proposed W-DSTAGNN method is its capacity to generate accurate and reliable long-term forecasts of traffic flows and prediction intervals for business deployment. Wavelets used within our framework act as catalysts to tackle the input time series's non-Gaussian and long-range dependence structures. This ensemble framework can be useful for other potential application areas, such as spatiotemporal predictions of epidemics or studying the evolving behavior of social networks. We tested our method using statistical tests to verify its robustness over benchmark models. Apart from point forecasts, our proposed model can generate interval forecasts that significantly contribute to the probability forecasting of traffic datasets. 

\section*{Code Availability Statement}
The source code for the W-DSTAGNN framework is made publicly available at \url{https://github.com/yash-jakhmola/w-dstagnn}. 

\bibliographystyle{ieeetr}
\bibliography{IEEEabrv,refs}

\begin{thebibliography}{10}

\bibitem{zheng2019deep}
Z.~Zheng, Y.~Yang, J.~Liu, H.-N. Dai, and Y.~Zhang, ``Deep and embedded
  learning approach for traffic flow prediction in urban informatics,'' {\em
  IEEE Transactions on Intelligent Transportation Systems}, vol.~20, no.~10,
  pp.~3927--3939, 2019.

\bibitem{vlahogianni2014short}
E.~I. Vlahogianni, M.~G. Karlaftis, and J.~C. Golias, ``Short-term traffic
  forecasting: Where we are and where we’re going,'' {\em Transportation
  Research Part C: Emerging Technologies}, vol.~43, pp.~3--19, 2014.

\bibitem{zhang2011data}
J.~Zhang, F.-Y. Wang, K.~Wang, W.-H. Lin, X.~Xu, and C.~Chen, ``Data-driven
  intelligent transportation systems: A survey,'' {\em IEEE Transactions on
  Intelligent Transportation Systems}, vol.~12, no.~4, pp.~1624--1639, 2011.

\bibitem{shahriari2020ensemble}
S.~Shahriari, M.~Ghasri, S.~Sisson, and T.~Rashidi, ``Ensemble of arima:
  combining parametric and bootstrapping technique for traffic flow
  prediction,'' {\em Transportmetrica A: Transport Science}, vol.~16, no.~3,
  pp.~1552--1573, 2020.

\bibitem{chandra2009predictions}
S.~R. Chandra and H.~Al-Deek, ``Predictions of freeway traffic speeds and
  volumes using vector autoregressive models,'' {\em Journal of Intelligent
  Transportation Systems}, vol.~13, no.~2, pp.~53--72, 2009.

\bibitem{castro2009online}
M.~Castro-Neto, Y.-S. Jeong, M.-K. Jeong, and L.~D. Han, ``Online-svr for
  short-term traffic flow prediction under typical and atypical traffic
  conditions,'' {\em Expert systems with applications}, vol.~36, no.~3,
  pp.~6164--6173, 2009.

\bibitem{lv2014traffic}
Y.~Lv, Y.~Duan, W.~Kang, Z.~Li, and F.-Y. Wang, ``Traffic flow prediction with
  big data: A deep learning approach,'' {\em Ieee transactions on intelligent
  transportation systems}, vol.~16, no.~2, pp.~865--873, 2014.

\bibitem{gao2022short}
Y.~Gao, C.~Zhou, J.~Rong, Y.~Wang, and S.~Liu, ``Short-term traffic speed
  forecasting using a deep learning method based on multitemporal traffic flow
  volume,'' {\em IEEE Access}, vol.~10, pp.~82384--82395, 2022.

\bibitem{ma2020multi}
Y.~Ma, Z.~Zhang, and A.~Ihler, ``Multi-lane short-term traffic forecasting with
  convolutional lstm network,'' {\em IEEE Access}, vol.~8, pp.~34629--34643,
  2020.

\bibitem{zhao2019deep}
W.~Zhao, Y.~Gao, T.~Ji, X.~Wan, F.~Ye, and G.~Bai, ``Deep temporal
  convolutional networks for short-term traffic flow forecasting,'' {\em Ieee
  Access}, vol.~7, pp.~114496--114507, 2019.

\bibitem{khan2023short}
A.~Khan, M.~M. Fouda, D.-T. Do, A.~Almaleh, and A.~U. Rahman, ``Short-term
  traffic prediction using deep learning long short-term memory: Taxonomy,
  applications, challenges, and future trends,'' {\em IEEE Access}, vol.~11,
  pp.~94371--94391, 2023.

\bibitem{berkani2023spatio}
S.~Berkani, B.~Guermah, M.~Zakroum, and M.~Ghogho, ``Spatio-temporal
  forecasting: A survey of data-driven models using exogenous data,'' {\em IEEE
  Access}, vol.~11, pp.~75191--75214, 2023.

\bibitem{hamdi2022spatiotemporal}
A.~Hamdi, K.~Shaban, A.~Erradi, A.~Mohamed, S.~K. Rumi, and F.~D. Salim,
  ``Spatiotemporal data mining: a survey on challenges and open problems,''
  {\em Artificial Intelligence Review}, pp.~1--48, 2022.

\bibitem{wang2020deep}
S.~Wang, J.~Cao, and S.~Y. Philip, ``Deep learning for spatio-temporal data
  mining: A survey,'' {\em IEEE transactions on knowledge and data
  engineering}, vol.~34, no.~8, pp.~3681--3700, 2020.

\bibitem{xie2020urban}
P.~Xie, T.~Li, J.~Liu, S.~Du, X.~Yang, and J.~Zhang, ``Urban flow prediction
  from spatiotemporal data using machine learning: A survey,'' {\em Information
  Fusion}, vol.~59, pp.~1--12, 2020.

\bibitem{traffic_planning}
Y.~E. Ay{\"o}zen and H.~{\.I}na{\c{c}}, ``Traffic planning in modern large
  cities paris and istanbul,'' {\em Scientific reports}, vol.~14, no.~1,
  pp.~1--10, 2024.

\bibitem{ermagun2018spatiotemporal}
A.~Ermagun and D.~Levinson, ``Spatiotemporal traffic forecasting: review and
  proposed directions,'' {\em Transport Reviews}, vol.~38, no.~6, pp.~786--814,
  2018.

\bibitem{shi2015infinite}
X.~Shi, H.~Feng, M.~Zhai, T.~Yang, and B.~Hu, ``Infinite impulse response graph
  filters in wireless sensor networks,'' {\em IEEE Signal Processing Letters},
  vol.~22, no.~8, pp.~1113--1117, 2015.

\bibitem{zhu2012graph}
X.~Zhu and M.~Rabbat, ``Graph spectral compressed sensing for sensor
  networks,'' in {\em 2012 IEEE International Conference on Acoustics, Speech
  and Signal Processing (ICASSP)}, pp.~2865--2868, IEEE, 2012.

\bibitem{mei2015signal}
J.~Mei and J.~M. Moura, ``Signal processing on graphs: Estimating the structure
  of a graph,'' in {\em 2015 IEEE International Conference on Acoustics, Speech
  and Signal Processing (ICASSP)}, pp.~5495--5499, IEEE, 2015.

\bibitem{ray2021optimized}
A.~Ray, T.~Chakraborty, and D.~Ghosh, ``Optimized ensemble deep learning
  framework for scalable forecasting of dynamics containing extreme events,''
  {\em Chaos: An Interdisciplinary Journal of Nonlinear Science}, vol.~31,
  no.~11, 2021.

\bibitem{choi2022graph}
J.~Choi, H.~Choi, J.~Hwang, and N.~Park, ``Graph neural controlled differential
  equations for traffic forecasting,'' in {\em Proceedings of the AAAI
  conference on artificial intelligence}, vol.~36, pp.~6367--6374, 2022.

\bibitem{dstagnn}
S.~Lan, Y.~Ma, W.~Huang, W.~Wang, H.~Yang, and P.~Li, ``Dstagnn: Dynamic
  spatial-temporal aware graph neural network for traffic flow forecasting,''
  in {\em International conference on machine learning}, pp.~11906--11917,
  PMLR, 2022.

\bibitem{yu2017spatio}
B.~Yu, H.~Yin, and Z.~Zhu, ``Spatio-temporal graph convolutional networks: A
  deep learning framework for traffic forecasting,'' {\em arXiv preprint
  arXiv:1709.04875}, 2017.

\bibitem{gman}
C.~Zheng, X.~Fan, C.~Wang, and J.~Qi, ``Gman: A graph multi-attention network
  for traffic prediction,'' in {\em Proceedings of the AAAI conference on
  artificial intelligence}, vol.~34, pp.~1234--1241, 2020.

\bibitem{pavlyuk2018spatiotemporal}
D.~Pavlyuk, ``Spatiotemporal big data challenges for traffic flow analysis,''
  in {\em Reliability and Statistics in Transportation and Communication:
  Selected Papers from the 17th International Conference on Reliability and
  Statistics in Transportation and Communication, RelStat’17, 18-21 October,
  2017, Riga, Latvia}, pp.~232--240, Springer, 2018.

\bibitem{fourier}
I.~N. Sneddon, {\em Fourier transforms}.
\newblock Courier Corporation, 1995.

\bibitem{fft}
W.~T. Cochran, J.~W. Cooley, D.~L. Favin, H.~D. Helms, R.~A. Kaenel, W.~W.
  Lang, G.~C. Maling, D.~E. Nelson, C.~M. Rader, and P.~D. Welch, ``What is the
  fast fourier transform?,'' {\em Proceedings of the IEEE}, vol.~55, no.~10,
  pp.~1664--1674, 1967.

\bibitem{waveletbook}
D.~B. Percival and A.~T. Walden, {\em Wavelet methods for time series
  analysis}, vol.~4.
\newblock Cambridge university press, 2000.

\bibitem{wtransformer}
L.~Sasal, T.~Chakraborty, and A.~Hadid, ``W-transformers: a wavelet-based
  transformer framework for univariate time series forecasting,'' in {\em 2022
  21st IEEE international conference on machine learning and applications
  (ICMLA)}, pp.~671--676, IEEE, 2022.

\bibitem{lee2021short}
K.~Lee, M.~Eo, E.~Jung, Y.~Yoon, and W.~Rhee, ``Short-term traffic prediction
  with deep neural networks: A survey,'' {\em IEEE Access}, vol.~9,
  pp.~54739--54756, 2021.

\bibitem{tedjopurnomo2020survey}
D.~A. Tedjopurnomo, Z.~Bao, B.~Zheng, F.~M. Choudhury, and A.~K. Qin, ``A
  survey on modern deep neural network for traffic prediction: Trends, methods
  and challenges,'' {\em IEEE Transactions on Knowledge and Data Engineering},
  vol.~34, no.~4, pp.~1544--1561, 2020.

\bibitem{hyndman2018forecasting}
R.~Hyndman, {\em Forecasting: principles and practice}.
\newblock OTexts, 2018.

\bibitem{petropoulos2022forecasting}
F.~Petropoulos, D.~Apiletti, V.~Assimakopoulos, M.~Z. Babai, D.~K. Barrow,
  S.~B. Taieb, C.~Bergmeir, R.~J. Bessa, J.~Bijak, J.~E. Boylan, {\em et~al.},
  ``Forecasting: theory and practice,'' {\em International Journal of
  Forecasting}, vol.~38, no.~3, pp.~705--871, 2022.

\bibitem{arima}
G.~E. Box, G.~M. Jenkins, G.~C. Reinsel, and G.~M. Ljung, {\em Time series
  analysis: forecasting and control}.
\newblock John Wiley \& Sons, 2015.

\bibitem{hong2011forecasting}
W.-C. Hong, Y.~Dong, F.~Zheng, and C.-Y. Lai, ``Forecasting urban traffic flow
  by svr with continuous aco,'' {\em Applied Mathematical Modelling}, vol.~35,
  no.~3, pp.~1282--1291, 2011.

\bibitem{chakraborty2019hybrid}
T.~Chakraborty, A.~K. Chakraborty, and Z.~Mansoor, ``A hybrid regression model
  for water quality prediction,'' {\em Opsearch}, vol.~56, pp.~1167--1178,
  2019.

\bibitem{yu2017deep}
R.~Yu, Y.~Li, C.~Shahabi, U.~Demiryurek, and Y.~Liu, ``Deep learning: A generic
  approach for extreme condition traffic forecasting,'' in {\em Proceedings of
  the 2017 SIAM international Conference on Data Mining}, pp.~777--785, SIAM,
  2017.

\bibitem{fclstm}
I.~Sutskever, O.~Vinyals, and Q.~V. Le, ``Sequence to sequence learning with
  neural networks,'' {\em Advances in neural information processing systems},
  vol.~27, 2014.

\bibitem{stresnet}
J.~Zhang, Y.~Zheng, D.~Qi, R.~Li, X.~Yi, and T.~Li, ``Predicting citywide crowd
  flows using deep spatio-temporal residual networks,'' {\em Artificial
  Intelligence}, vol.~259, pp.~147--166, 2018.

\bibitem{gcn}
J.~Bruna, W.~Zaremba, A.~Szlam, and Y.~LeCun, ``Spectral networks and locally
  connected networks on graphs,'' {\em arXiv preprint arXiv:1312.6203}, 2013.

\bibitem{chebnet}
M.~Defferrard, X.~Bresson, and P.~Vandergheynst, ``Convolutional neural
  networks on graphs with fast localized spectral filtering,'' {\em Advances in
  neural information processing systems}, vol.~29, 2016.

\bibitem{dcrnn}
Y.~Li, R.~Yu, C.~Shahabi, and Y.~Liu, ``Diffusion convolutional recurrent
  neural network: Data-driven traffic forecasting,'' {\em arXiv preprint
  arXiv:1707.01926}, 2017.

\bibitem{stsgcn}
C.~Song, Y.~Lin, S.~Guo, and H.~Wan, ``Spatial-temporal synchronous graph
  convolutional networks: A new framework for spatial-temporal network data
  forecasting,'' in {\em Proceedings of the AAAI conference on artificial
  intelligence}, vol.~34, pp.~914--921, 2020.

\bibitem{graphwavenet}
Z.~Wu, S.~Pan, G.~Long, J.~Jiang, and C.~Zhang, ``Graph wavenet for deep
  spatial-temporal graph modeling,'' {\em arXiv preprint arXiv:1906.00121},
  2019.

\bibitem{agcrn}
L.~Bai, L.~Yao, C.~Li, X.~Wang, and C.~Wang, ``Adaptive graph convolutional
  recurrent network for traffic forecasting,'' {\em Advances in neural
  information processing systems}, vol.~33, pp.~17804--17815, 2020.

\bibitem{joo2015time}
T.~W. Joo and S.~B. Kim, ``Time series forecasting based on wavelet
  filtering,'' {\em Expert Systems with Applications}, vol.~42, no.~8,
  pp.~3868--3874, 2015.

\bibitem{liu2017time}
T.~Liu, H.~Wei, C.~Zhang, and K.~Zhang, ``Time series forecasting based on
  wavelet decomposition and feature extraction,'' {\em Neural Computing and
  Applications}, vol.~28, pp.~183--195, 2017.

\bibitem{panja2023ensemble}
M.~Panja, T.~Chakraborty, S.~S. Nadim, I.~Ghosh, U.~Kumar, and N.~Liu, ``An
  ensemble neural network approach to forecast dengue outbreak based on
  climatic condition,'' {\em Chaos, Solitons \& Fractals}, vol.~167, p.~113124,
  2023.

\bibitem{sengupta2023forecasting}
S.~Sengupta, T.~Chakraborty, and S.~K. Singh, ``Forecasting cpi inflation under
  economic policy and geo-political uncertainties,'' {\em arXiv preprint
  arXiv:2401.00249}, 2023.

\bibitem{borah2024wavecatboost}
J.~Borah, T.~Chakraborty, M.~S.~M. Nadzir, M.~G. Cayetano, and S.~Majumdar,
  ``Wavecatboost for probabilistic forecasting of regional air quality data,''
  {\em arXiv preprint arXiv:2404.05482}, 2024.

\bibitem{grinsted2004application}
A.~Grinsted, J.~C. Moore, and S.~Jevrejeva, ``Application of the cross wavelet
  transform and wavelet coherence to geophysical time series,'' {\em Nonlinear
  processes in geophysics}, vol.~11, no.~5/6, pp.~561--566, 2004.

\bibitem{jiang2005dynamic}
X.~Jiang and H.~Adeli, ``Dynamic wavelet neural network model for traffic flow
  forecasting,'' {\em Journal of transportation engineering}, vol.~131, no.~10,
  pp.~771--779, 2005.

\bibitem{xie2007short}
Y.~Xie, Y.~Zhang, and Z.~Ye, ``Short-term traffic volume forecasting using
  kalman filter with discrete wavelet decomposition,'' {\em Computer-Aided
  Civil and Infrastructure Engineering}, vol.~22, no.~5, pp.~326--334, 2007.

\bibitem{xiao2003fuzzy}
H.~Xiao, H.~Sun, B.~Ran, and Y.~Oh, ``Fuzzy-neural network traffic prediction
  framework with wavelet decomposition,'' {\em Transportation research record},
  vol.~1836, no.~1, pp.~16--20, 2003.

\bibitem{sun2015novel}
Y.~Sun, B.~Leng, and W.~Guan, ``A novel wavelet-svm short-time passenger flow
  prediction in beijing subway system,'' {\em Neurocomputing}, vol.~166,
  pp.~109--121, 2015.

\bibitem{zhang2019wavelet}
N.~Zhang, X.~Guan, J.~Cao, X.~Wang, and H.~Wu, ``Wavelet-hst: A wavelet-based
  higher-order spatio-temporal framework for urban traffic speed prediction,''
  {\em IEEE Access}, vol.~7, pp.~118446--118458, 2019.

\bibitem{hydrology}
Y.-F. Sang, ``A review on the applications of wavelet transform in hydrology
  time series analysis,'' {\em Atmospheric research}, vol.~122, pp.~8--15,
  2013.

\bibitem{epidemics}
M.~Panja, T.~Chakraborty, U.~Kumar, and N.~Liu, ``Epicasting: an ensemble
  wavelet neural network for forecasting epidemics,'' {\em Neural Networks},
  vol.~165, pp.~185--212, 2023.

\bibitem{subtidal}
D.~B. Percival and H.~O. Mofjeld, ``Analysis of subtidal coastal sea level
  fluctuations using wavelets,'' {\em Journal of the American Statistical
  Association}, vol.~92, no.~439, pp.~868--880, 1997.

\bibitem{statsmodel}
S.~Seabold and J.~Perktold, ``Statsmodels: econometric and statistical modeling
  with python.,'' {\em SciPy}, vol.~7, p.~1, 2010.

\bibitem{svr}
H.~Drucker, C.~J. Burges, L.~Kaufman, A.~Smola, and V.~Vapnik, ``Support vector
  regression machines,'' {\em Advances in neural information processing
  systems}, vol.~9, 1996.

\bibitem{var}
J.~D. Hamilton, {\em Time series analysis}.
\newblock Princeton university press, 2020.

\bibitem{stgcn}
B.~Yu, H.~Yin, and Z.~Zhu, ``Spatio-temporal graph convolutional networks: A
  deep learning framework for traffic forecasting,'' {\em arXiv preprint
  arXiv:1709.04875}, 2017.

\bibitem{mcb}
A.~J. Koning, P.~H. Franses, M.~Hibon, and H.~O. Stekler, ``The m3 competition:
  Statistical tests of the results,'' {\em International journal of
  forecasting}, vol.~21, no.~3, pp.~397--409, 2005.

\bibitem{cp_book}
V.~Vovk, A.~Gammerman, and G.~Shafer, {\em Algorithmic learning in a random
  world}, vol.~29.
\newblock Springer, 2005.

\end{thebibliography}

\vfill

\end{document}